\documentclass[acmtog, nonacm]{acmart}
\acmSubmissionID{742}

\usepackage{booktabs} 

\usepackage[ruled]{algorithm2e} 

\SetAlFnt{\small}
\SetAlCapFnt{\small}
\SetAlCapNameFnt{\small}
\SetAlCapHSkip{0pt}

\usepackage{listings} 
\usepackage{multirow} 
\usepackage{subcaption} 
\usepackage{lipsum}  

\setcopyright{rightsretained}

\begin{document}
\title{Enhancing Diffusion Models with 3D Perspective Geometry Constraints}

\author{Rishi Upadhyay}
\email{rishiu@ucla.edu}
\affiliation{%
  \institution{University of California, Los Angeles}
  \country{USA}
}

\author{Howard Zhang}
\email{hwd15508@ucla.edu}
\affiliation{%
  \institution{University of California, Los Angeles}
    \country{USA}
}

\author{Yunhao Ba}
\email{yhba@ucla.edu}
\affiliation{%
  \institution{University of California, Los Angeles}
  \country{USA}
}
\affiliation{%
  \institution{Sony AI}
    \country{USA}
}

\author{Ethan Yang}
\email{eyang657@ucla.edu}
\affiliation{%
  \institution{University of California, Los Angeles}
    \country{USA}
}

\author{Blake Gella}
\email{bgella118827@ucla.edu}
\affiliation{%
  \institution{University of California, Los Angeles}
    \country{USA}
}

\author{Sicheng Jiang}
\email{sicheng2020@ucla.edu}
\affiliation{%
  \institution{University of California, Los Angeles}
    \country{USA}
}

\author{Alex Wong}
\email{alex.wong@yale.edu}
\affiliation{%
  \institution{Yale University}
    \country{USA}
}

\author{Achuta Kadambi}
\email{achuta@ee.ucla.edu}
\affiliation{%
  \institution{University of California, Los Angeles}
    \country{USA}
}

\begin{abstract}
While perspective is a well-studied topic in art, it is generally taken for granted in images. However, for the recent wave of high-quality image synthesis methods such as latent diffusion models, perspective accuracy is not an explicit requirement. Since these methods are capable of outputting a wide gamut of possible images, it is difficult for these synthesized images to adhere to the principles of linear perspective. We introduce a novel geometric constraint in the training process of generative models to enforce perspective accuracy. We show that outputs of models trained with this constraint both appear more realistic and improve performance of downstream models trained on generated images. Subjective human trials show that images generated with latent diffusion models trained with our constraint are preferred over images from the Stable Diffusion V2 model ~70\% of the time. SOTA monocular depth estimation models such as DPT and PixelFormer, fine-tuned on our images, outperform the original models trained on real images by up to 7.03\% in RMSE and 19.3\% in SqRel on the KITTI test set for zero-shot transfer.
\end{abstract}

%
%


\newcommand{\heightTeaser}{.29\textwidth}
\begin{teaserfigure}
  \centering
  \includegraphics[width=\textwidth]{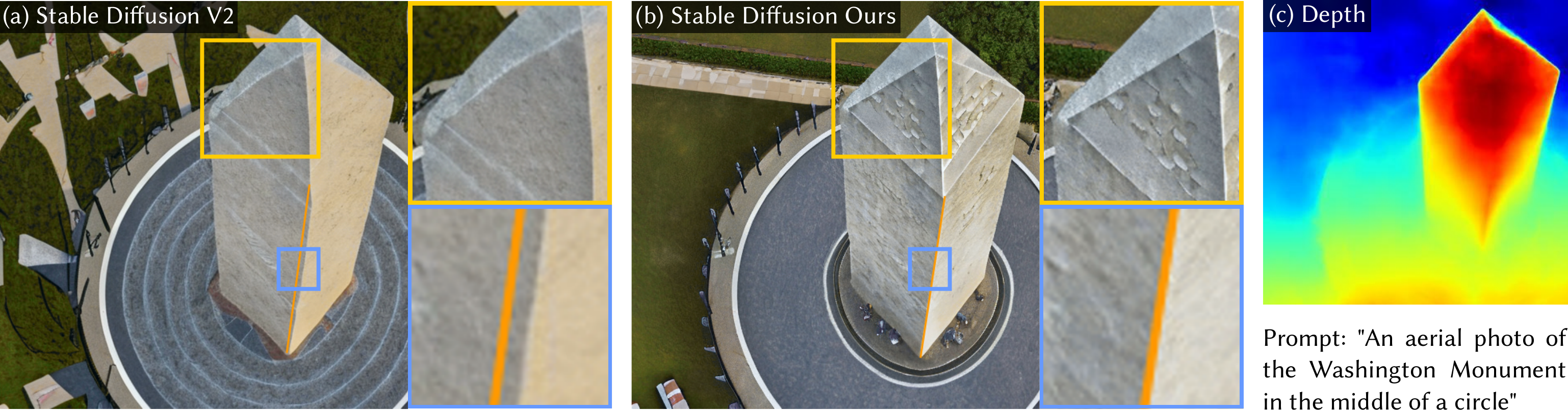}
  \caption{\textbf{Images generated with our novel geometric constraint preserve straight lines and perspective.} (a) An image generated by stable diffusion v2, (b) An image generated by our fine-tuned diffusion model, and (c) the depth map and prompt both models were conditioned on.}
  \label{fig:teaser}
\end{teaserfigure}

\maketitle

\section{Introduction}
"Re-draw The School of Athens in the style of Van Gogh", "Show an aerial viewpoint of the Washington Monument". The introduction of recent text-to-image synthesis methods such as latent diffusion models has drastically increased our creative capabilities. These models can generate anything from a Renaissance style painting to an everyday smartphone selfie from just a simple text prompt. However, as powerful as these models can be, their limited ability to adhere to physical constraints that are explicitly present in natural images restricts their potential~\cite{wang2022synthetic}. In contrast, traditional methods of image generation such as hand-drawn art or ray-traced images place careful attention on ensuring an accurate physical environment including geometry and lighting. 

One of the largest advancements in the photo-realism of hand-drawn art was the development of a system to draw accurate perspective geometry in the 1400s. While the gap between real and generated images is not as large for diffusion models as it was back then, a greater consideration for perspective accuracy can have a similarly large impact in the photo-realism of their outputs. 

Perspective is one of the most important physical constraints because it ensures object properties such as size, relative location, and depth are accurately represented. In a sense, it ensures physical accuracy~\cite{kadambi2020blending}. This allows the use of perspective accurate data for downstream tasks such as camera calibration~\cite{beardsley1992camera, caprile1990using, chen19913, he2007novel, li2010simultaneous}, 3D reconstruction~\cite{guillou2000using, Wang2009WhatCW}, scene understanding~\cite{Han2009Bottom, geiger20143D, satkin2012data}, and SLAM~\cite{Camposeco2015Using, georgis2022vp, Lim2022UV}.

However, current diffusion based image generators such as~\cite{rombach2022high, razavi2019generating, radford2021learning, bau2021paint, yu2022scaling} do not generate perspectively accurate data~\cite{farid_arxiv_perspective}. Please refer to Fig.~\ref{fig:teaser}(a) for an example of this phenomenon. This is because latent diffusion models typically lack the interpretability necessary for explicit encoding of a physical prior such as perspective in the model architecture~\cite{kadambi2023incorporating}. By utilizing a novel loss function that ensures the gradient field of an image aligns with its expected vanishing points, we are able to encode this physical prior. By enforcing this perspective prior on generated images, we also increase the accuracy of object properties important for downstream computer vision tasks and photo-realism. 

As it turns out, the perspective correctness of an image has a strong influence over its overall scene coherence and therefore realism. This is most likely true because, as mentioned before, perspective provides crucial information regarding the size, relative location, and depth of a scene. To illustrate this, we set up a human subjective test where the photo-realism of our perspective-corrected images is put to the test. We show that latent diffusion models which utilize our novel perspective loss generate images that are rated as more realistic an overwhelming majority of the time as compared to images generated by the base diffusion model. We also verify the visual benefits of our proposed constraint by applying it to the inpainting task. We show that inpainted images generated from models trained with our loss consistently appear more perceptually similar to the original image than images from models without our loss.

Additionally, images generated with our novel loss prove beneficial to the accuracy of downstream tasks which are inherently reliant on these same object properties. As proof of this concept, we fine-tune multiple SOTA monocular depth estimation models such as DPT~\cite{ranftl2021vision} and PixelFormer~\cite{agarwal2023attention}. We show that training on data with accurate perspective leads to models with higher performance that can capture high-frequency details to a higher degree. 

\subsection{Contributions}
In summary, we make the following contributions: 
\begin{itemize}
    \item We introduce a novel geometric constraint on the training process of latent diffusion models to enforce perspective accuracy.
    \item We show that images from models trained with this constraint appear more realistic than models trained without this constraint 69.6\% of the time.
    \item We demonstrate that downstream tasks which benefit from more geometrically accurate inputs (such as monocular depth estimation) improve by up to 7.03\% in RMSE and 19.3\% in SqRel.
\end{itemize}
\vfill

\section{Related Work} 

\subsection{Synthetic Image Generation}

Image generation, while a popular task, has proven to be difficult because of the high dimensional space and variety of images. One of the most popular techniques for image generation has been Generative Adversarial Networks (GANs)~\cite{GANs}. While GANs are capable of high quality image synthesis~\cite{brock2018large}, they are limited by the fact that they are difficult to train, often failing to converge or collapsing into a mode where all generated images are the same~\cite{pmlr-v80-mescheder18a, pmlr-v70-arjovsky17a}. Another popular image generation technique is Variational Auto-encoders (VAEs)~\cite{iclr_KingmaW13} which have stronger theoretical guarantees, but cannot match GANs in image quality~\cite{child2021very, vahdat2020NVAE}. Recently, diffusion models~\cite{sohl2015deep} for image generation have grown in popularity. These models work by reversing a diffusion process which adds noise to high quality images and are capable of generating high quality samples from a variety of distributions~\cite{ho2020denoising, dhariwal2021diffusion, daras2022soft}. Subsequent works have expanded the scope even further by adding text guidance to the diffusion process~\cite{ramesh2022hierarchical, saharia2022photorealistic}, simplifying the inverse process~\cite{wallace2022edict}, and reformulating the diffusion process to occur in a latent space for speed benefits~\cite{rombach2022high}. While recent work has explored guiding diffusion models in various ways~\cite{wallace2023end, ho2022classifier, meng2023distillation, rombach2022text}, most diffusion models rely almost entirely on their vast datasets and text encoders for priors on scene composition and object properties. This means that there are no explicit guarantees that generated images will be physically accurate, making them a poor fit for use in synthetic datasets. Our work aims to add 3D geometry constraints to image generators in order to improve the quality of generated images.

A specific task in the space of synthetic image generation that is related to our work is the edge-to-image synthesis problem. In this task, the diffusion model is conditioned on both a text prompt as well as an edge map of the scene we want to generate~\cite{batzolis2022non, batzolis2021conditional}. Although this is similar to our task in terms of constraints on edges in an image, they are not quite the same problem: for the edge-to-image task, the goal of training is to have a model which can follow the provided edge map faithfully~\cite{xu2020e2i}. If this is achieved, perspective accuracy can be achieved by providing perspectively accurate edge maps. However, for our work, the task is to instead train a model that can generate perspectively accurate images without access to an edge map, meaning our models require less input and are more general.

\begin{figure*}
    \newcommand{\widthfig}{.326\textwidth}
    \centering
  \includegraphics[width=\widthfig]{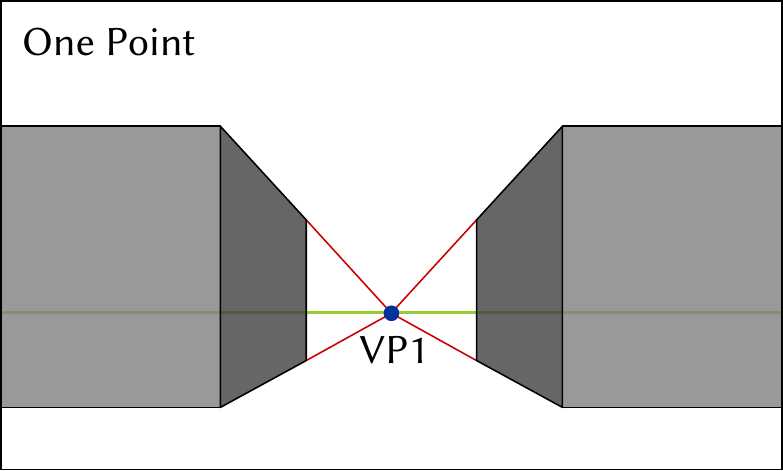}
  \hfill
  \includegraphics[width=\widthfig]{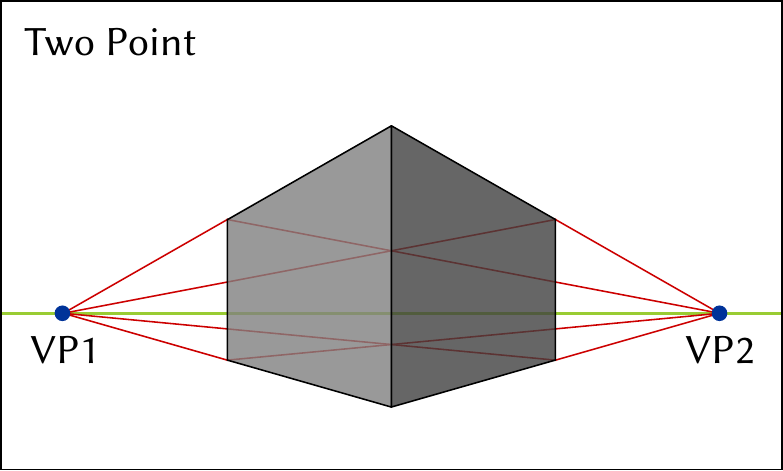}
  \hfill
  \includegraphics[width=\widthfig]{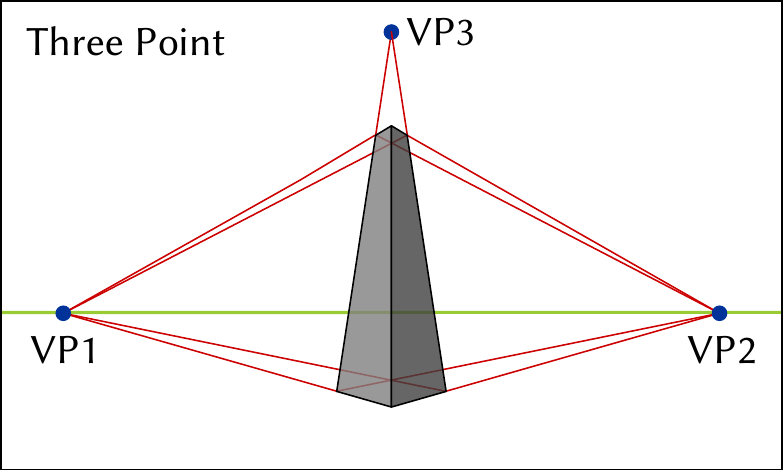}
  \caption{\textbf{Examples of one, two, and three-point linear perspective.} Vanishing points are labeled in blue, perspective lines are in red, and the horizon lines are in light green. One-point perspective is typically used when there is one focal point of the image or when only one side of an object is visible. Two-point perspective is used to illustrate multiple sides of an object, while three-point perspective is used for viewpoints that are above or below the horizon line of the 3D scene.}
  \label{fig:perspective_explain}
\end{figure*}

\subsection{Vanishing Points in Computer Vision}

Vanishing points have many varied and important uses in computer vision. One common use for vanishing points is camera calibration. Early examples of this include~\cite{caprile1990using, beardsley1992camera, chen19913} who use vanishing point geometry to compute the intrinsics and extrinsics of one or more cameras given single or multiple images. Subsequent papers, such as~\cite{he2007novel, li2010simultaneous}, provided improved techniques that were simpler or required less data and assumptions. In addition, newer works began to not only compute camera parameters, but also use them to compute 3D reconstructions of single images~\cite{guillou2000using, Wang2009WhatCW}. Beyond camera calibration, vanishing points are also useful for general scene understanding.~\cite{Han2009Bottom} use vanishing points to help create generative grammar for synthetic scenes,~\cite{geiger20143D} use vanishing points as priors for 3D scene and traffic understanding, and~\cite{satkin2012data} estimate 3D models from singular images using vanishing point priors. Vanishing points are also particularly useful for road detection thanks to easily identifiable perspective lines, as demonstrated by~\cite{liou1987road, kong2009vanishing}. Vanishing points are also regularly used in SLAM techniques. \cite{Lee2009VPass} were one of the first in this space, using vanishing points to identify the heading of a robot for navigation. Subsequent works further expanded the capabilities of SLAM systems built on vanishing points including~\cite{Camposeco2015Using, Lim2022UV, georgis2022vp} who use vanishing points to identify direction and perform structural mapping of scenes in real-time. Given the significance of vanishing points in computer vision, we aim to enhance image generators with accurate perspective, in order to benefit photo-realism and downstream tasks. 

In additional to vanishing points, perspective has been used in computer vision for computational photography tasks. For example, many works use perspective principles to allow for editing the focal length and camera position of an image after it is taken~\cite{Badki17, 10.1111:cgf.14458}. Another application of perspective are techniques which aim to reduce distortion in wide-angle images~\cite{10.1145/1576246.1531349, shih2019distortion}. These techniques often learn the perspective projection of an image and then find transformations to achieve the desired un-distorted images. Other works have also gone the opposite direction by introducing new types of perspective projections that are not necessarily physically accurate but can result in artistic and aesthetic images~\cite{cubist, 10.1007/978-3-7091-6303-0_12}.

\subsection{Monocular Depth Estimation}

Supervised methods for monocular depth estimation typically require paired image and depth data. One of the first works in this area was Make3D~\cite{saxena2008make3d} which relied on hand-crafted features and Markov random fields. Subsequent works then applied deep learning to the problem, starting with multi-scale convolutional networks~\cite{eigen2014depth} and followed by conditional random fields~\cite{Li2015DepthAS}, residual networks~\cite{laina2016deeper}, convolutional neural fields~\cite{liu2015learning, xu2018structured}, and most recently transformers~\cite{ranftl2021vision, ranftl2020towards, agarwal2023attention}. Many approaches also take advantage of known geometric relationships, such as normals~\cite{qi2018geonet} and planes~\cite{lee2019big, yang2018recovering}. Newer techniques have also taken an unsupervised approach~\cite{wong2019bilateral, fei2019geo} or use multi-modal data capture~\cite{singh2023depth}. However, most supervised monocular depth estimation models are limited by the availability of paired data on which to train as this data is difficult to collect. 

In order to overcome the challenge of a lack of sufficient training data, many techniques turn to synthetic datasets. The renderers used to generate the images in these datasets can often generate corresponding ground-truth data, making it simple to acquire pixel-aligned ground-truth depth maps. In addition, these renderers often allow for different types of data, such as varied weather conditions or indoor vs. outdoor scenes, making them an attractive way to get training data. Examples of such datasets include Virtual KITTI, a photorealistic copy of the popular self-driving dataset KITTI~\cite{gaidon2016virtual, Geiger2013IJRR} and SYNTHIA, a dataset that includes depth and semantic segmentation information for images of a synthetic city~\cite{ros2016synthia, bengar2019temporal}. Although these datasets are often quite realistic, there are often key differences between synthetic and real images which leads models trained on synthetic images to achieve lower performance when tested on real datasets compared to models trained and tested on real images. This difference in performance is referred to as the Sim2Real gap. As monocular depth estimation is a popular task, many works have attempted to address the problem of the Sim2Real gap~\cite{cao2018dida, damodaran2018deepjdot, long2015learning, rozantsev2018beyond, sankaranarayanan2018generate}. However, all of these techniques approach the problem by attempting to improve the neural network architectures. On the other hand, we approach this problem from the perspective of improving the synthetic data used to train the neural networks.

In addition to monocular depth estimation, the techniques we describe in the paper can be easily applied to the task of depth completion as well since the data format is the same for both tasks~\cite{wong2021unsupervised, 9918022, wong2021learning}.

\section{Perspective Background}

\subsection{Linear Perspective}
\label{perspinart}
Although perspective is a word commonly used in a variety of contexts, it has a very specific meaning in terms of art and photography: techniques used to draw objects in 2D such that their 3D attributes are correctly modeled. In practice, perspective refers to a multitude of different techniques which can be used to create a 3D feel, but the most common technique is called linear perspective. In linear perspective, all mutually parallel lines, on the same or parallel planes, in 3D space, converge to a single point in the image plane which is referred to as a vanishing point. The only exception to this rule is sets of lines that are exactly parallel to the camera sensor. In this case, these lines are also parallel in the image plane. A typical drawing/image often has anywhere from one to three vanishing points, with the number of vanishing points determining the style and view of the drawing/image. Another key component of linear perspective is the horizon line. The horizon line is a horizontal line that represents the viewer's eye level in an image, and typically at least one of the vanishing points of an image lies on this line. A visualization of these principles can be found in Fig.~\ref{fig:perspective_explain}. 

\subsection{Perspective Consistency in Images}
Perspective in images is not always easy to confirm, as the vanishing points of an image can only be easily identified with the aid of parallel lines in 3D space, which may not always exist in images. For images that do have sets of parallel lines, perspective consistency can be verified by extending sets of parallel lines in either direction until they intersect and ensuring that all pairs of lines in a set intersect at the same point.

\paragraph{Natural Images} By the math of perspective projection for a pinhole camera, an point $\mathbf{X} = (X,Y,Z)$ is projected to a point $\mathbf{x} = (x,y) = (fX/Z, fY/Z)$~\cite{10.5555/971144}. If we are concerned with a line $L = \mathbf{O} + t \mathbf{D}$, after replacing $X, Y, Z$ above with the equation for a line and taking the limit as $t$ goes to positive/negative infinity, we see that the final projected point is dependent on only $\mathbf{D}$. Therefore, sets of parallel lines, that are not parallel to the camera plane, will all come together to the same point, known  as a vanishing point.

\paragraph{Synthetic Images} Although natural images are forced to follow perspective rules, there are no such restrictions on synthetic images, particularly images generated by deep learning approaches. Most of the loss functions used to train these models are focused on image quality or how well prompts are followed, meaning physical properties such as perspective, shadows, or lighting can often be neglected~\cite{farid_arxiv_perspective, farid_arxiv_lighting}. An example of this can be seen in Fig.~\ref{fig:teaser}(a).

\label{sec:persp_loss_func}
\begin{figure}[t]
    \centering
    {\includegraphics[width=.45\textwidth]{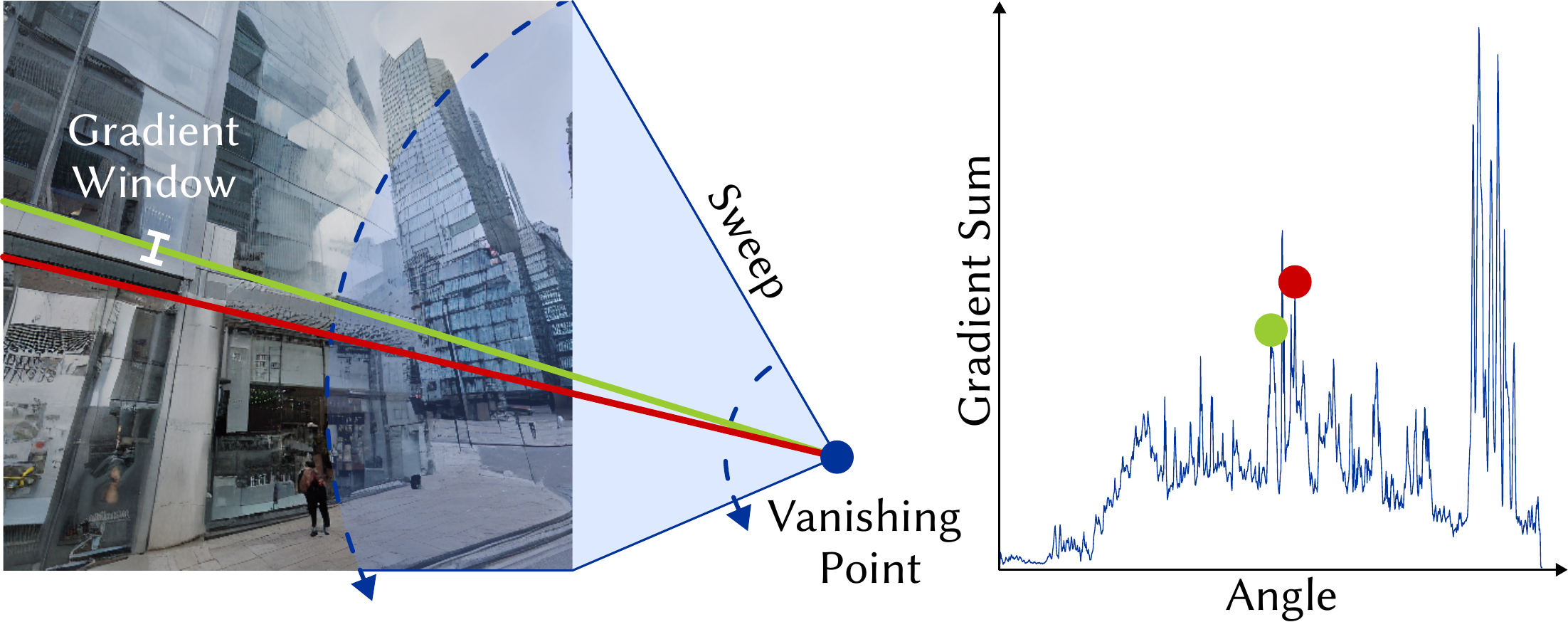}}
    \caption{\textbf{Graphical description of our geometric constraint.} \textit{Left:} A visualization of how the loss function sweeps lines across the image. \textit{Right:} $D(v, \textbf{x})$ plotted for the image at right. The red and yellow lines in the left plot are identified by the corresponding dots. }
    \label{fig:loss_function}
\end{figure}

\section{Improving perspective accuracy of generated images}
Our fine-tuned model is built on top of the latent diffusion models introduced by~\cite{rombach2022high}, using code from~\cite{pinkney2022repo}. We describe the latent diffusion process in Section~\ref{sec:latent_diff_models}. We add a new term to the traditional loss function and train on a specialized dataset that provides ground truth vanishing points. This new constraint is described in Section~\ref{sec:persp_loss_func}.

\subsection{Latent Diffusion Models}
\label{sec:latent_diff_models}
Traditional image generation diffusion models are concerned with a forward diffusion process over images $\mathbf{x}_0$,...,$\mathbf{x}_T$:
\begin{equation}
q(\mathbf{x}_t|\mathbf{x}_{t-1}) =  \mathcal{N}(\sqrt{\alpha_t} \mathbf{x}_{t-1}, (1 - \alpha_t) \mathbf{I}),
\end{equation}
where $q$ is the forward diffusion function, $t$ is the current time step, and $\mathbf{I}$ is the identity. $\alpha_t = 1 - \beta_t$ and $\beta_1$,...,$\beta_T$ compose a pre-selected variance schedule. The reverse process is then parameterized as:
\begin{equation}
p(\mathbf{x}_{t-1} | \mathbf{x}_t) = \mathcal{N}(\mathbf{\mu}_\theta(\mathbf{x}_t, t), \mathbf{\Sigma}(\mathbf{x}_t, t)),
\end{equation}
where $p$ is defined as the reverse diffusion function and $\mathbf{\Sigma}(\mathbf{x}_t, t)$ is typically set to time-dependent constants. $\mathbf{\mu}_\theta(\mathbf{x}_t, t)$ is defined as:
\begin{equation}
    \mathbf{\mu}_\theta(\mathbf{x}_t, t) = \frac{1}{\sqrt{\alpha_t}} \left( \mathbf{x}_t - \frac{\beta_t}{\sqrt{1 - \overline{\alpha}_t}} \mathbf{\epsilon}_\theta (\mathbf{x}_t, t) \right),
\end{equation}
where $\overline{\alpha}_t = \Pi_{i=1}^t \alpha_i$, and $\epsilon_\theta (\mathbf{x}_t, t)$ is a learned function parameterized by a UNet model~\cite{ronneberger2015u} with learned parameters $\theta$. Based on this, the traditional diffusion model loss is as follows:
\begin{equation}
   L_{\text{DM}} = \mathbb{E}_{\mathbf{x},\epsilon \sim \mathbb{N}(0,1), t} \left[ {\| \epsilon - \epsilon_\theta (\mathbf{x}_t, t) \|}^2_2 \right].
\end{equation}
More details and derivations can be found in~\cite{ho2020denoising}. Latent diffusion models work very similarly, but perform the forward and reverse diffusion processes in latent spaces. Specifically, an encoder and decoder are introduced to translate to and from the latent space. The encoder is defined as: $\mathcal{E}: X \in R^{H \times W \times 3} \mapsto Z \in R^{h \times w \times 3}$, while the decoder is defined as: $\mathcal{D}:Z \in R^{h \times w \times 3} \mapsto X \in R^{H \times W \times 3}$, where $h = H / f$, $w = W / f$ and $f$ is a downsampling factor. With this formulation, the loss function now becomes:
\begin{equation}
   L_{\text{LDM}} = \mathbb{E}_{\mathcal{E}(\mathbf{x}),\epsilon \sim \mathbb{N}(0,1), t} \left[ {\| \epsilon - \epsilon_\theta (z_t, t) \|}^2_2 \right],
\end{equation}
where the image $x_t$ is replaced by its latent space representation $z_t$.

\SetKwProg{Fn}{Function}{}{end}
\begin{algorithm}
\caption{Algorithm to compute perspective loss}\label{algo:persploss}
\DontPrintSemicolon
\Fn{perspective\_loss$(\mathbf{x}, \hat{\mathbf{x}}, \mathbf{v}_{\mathbf{x}})$}{
\SetKwInOut{Input}{Input}\SetKwInOut{Output}{Output}
\Input{Image $\hat{\mathbf{x}}$}
\Input{Ground Truth image $\mathbf{x}$}
\Input{Vanishing Points $\mathbf{v}_{\mathbf{x}}$}
\BlankLine
$G_\mathbf{x} \gets$ img\_derivative($\mathbf{x}$) \;
$G_{\hat{\mathbf{x}}} \gets$ img\_derivative($\hat{\mathbf{x}})$ \;
$loss \gets 0.0$ \;
\ForEach{$v \in \mathbf{v}_{\mathbf{x}}$}{
    $\phi_{\text{min}}, \phi_{\text{max}} \gets$ calc\_image\_angle($v$) \;
    \For{$i \gets 0; i < N; i = i+1$}{
        $angle \gets \frac{i}{N} * (\phi_{\text{max}} - \phi_{\text{min}}) +  \phi_{\text{min}} $ \;
        $d \gets$ calc\_perp\_vec($angle$) \;
        $p \gets$ get\_line\_pixels($v$, $angle$) \;
        $D(i) \gets \sum_p |G_{\hat{\mathbf{x}}} \cdot d|$ \;
        $D_{\text{gt}}(i) \gets \sum_p |G_{\mathbf{x}} \cdot d|$ \;
    $loss \gets loss + \text{norm}(D - D_{\text{gt}})$ \;
    }
$loss \gets loss / |\mathbf{v}|$ \;
}
\Return $loss$
}
\end{algorithm}

 In order to add perspective priors to a latent diffusion model, we add an additional perspective loss term. At a high level, this loss works by sweeping lines extending out from a vanishing point over the image and calculating the sum of image gradients across the line, as illustrated in Fig.~\ref{fig:loss_function}. Pseudocode for this algorithm is shown in Alg.~\ref{algo:persploss}. This sum is designed to represent how "edge-like" the region along that line is in the image. We can then write our new loss as:
\begin{equation}
\label{eq:new-loss}
\begin{split}
L_{\text{DM}} = \mathbb{E}_{\mathcal{E}(\mathbf{x}),\epsilon \sim \mathbb{N}(0,1), t} \left[ {\| \epsilon - \epsilon_\theta (z_t, t) \|}^2_2 \right] + \\  \lambda \mathbb{E}_{\mathbf{x},\epsilon \sim \mathbb{N}(0,1), v} \left[ L_{\text{persp}}(\hat{\mathbf{x}}, \mathbf{x}, \mathbf{v}_{\mathbf{x}}) \right].
\end{split}
\end{equation}
where $\lambda$ is a weight factor for our perspective loss, $\mathbf{v}_{\mathbf{x}}$ is a set of vanishing points in image space and $\hat{\mathbf{x}}$ is our reconstructed image, which can be written as:
\begin{equation}
\hat{\mathbf{x}} = \mathcal{D}\left(\frac{1}{\sqrt{\overline{\alpha}_t}} \left(\mathbf{z}_t - \sqrt{1 - \overline{\alpha}_t} \mathbf{\epsilon_\theta}(z_t, t) \right)\right).
\end{equation}
where $t$ is randomly chosen between 0 and $T$ for each iteration. In order to define $L_{\text{persp}}$, we first define some intermediate quantities: 
\begin{itemize}
    \item $G_{\mathbf{x}}$ represents the gradients of an image $\mathbf{x}$ computed with a 3x3 Sobel filter.
    \item $\phi_\text{min}$ and $\phi_\text{max}$ represent the minimum and maximum angle from the vanishing point to a corner of the image relative to the x-axis.
    \item $\phi_0$,...,$\phi_n$ represent $n$ equally-spaced angles between $\phi_\text{min}$ and $\phi_\text{max}$.
    \item $v$ represents a particular vanishing point in the set $\mathbf{v}_{\mathbf{x}}$.
    \item $l_i(v, k)$ represents a point at time $k$ on a ray $l_i(v)$ starting at $v$ in the direction of $\phi_i$.
    \item $d_i(v)$ represents a vector perpendicular to the line $l_i(v)$.
\end{itemize}
Using these, we define:
\begin{equation} \label{eq:8}
D_i(v, \mathbf{x}) = \int_{k_0}^{k_1} \lvert d_i \cdot G_{\mathbf{x}}(l_i(v, k)) \rvert  dk,
\end{equation}
where $k_0$ and $k_1$ represent the times of the intersection of $l_i(v)$ with $\mathbf{x}$. $D_i(v,\mathbf{x})$ is then our measure of how "edge-like" the region along this ray is, and we can then define:
\begin{equation}
L_{\text{persp}}(\hat{\mathbf{x}}, \mathbf{x}, \mathbf{v}_{\mathbf{x}}) = \frac{1}{|\mathbf{v}_{\mathbf{x}}|} \sum_{v \in \mathbf{v}_{\mathbf{x}}} || D(v, \hat{\mathbf{x}}) - D(v, {\mathbf{x}}) ||_2\\.
\end{equation}
In practice, the integral in Eq.~\ref{eq:8} becomes a sum over the image pixels that the line intersects. Our loss function was implemented entirely in PyTorch and is fully differentiable end-to-end.

\section{Experiments}

In order to evaluate our proposed constraint, we conduct comprehensive experiments. In Section~\ref{ssc:latent-diffusion-training}, we detail how we fine-tune latent diffusion models with the proposed constraint, in Section~\ref{ssc:depth-est-training}, we detail how we fine-tune monocular depth estimation models on images generated from our fine-tuned models. In Section~\ref{ssc:human-subj-methods}, we describe how we evaluate the photo-realism of images generated from our fine-tuned models, and in Section~\ref{ssc:ablation-methods}, we describe our ablation studies.

\subsection{Training Latent Diffusion Models}
\label{ssc:latent-diffusion-training}

For all of our image generation experiments, we build off the depth-conditioned Stable Diffusion V2 model from~\cite{rombach2022high}. This model is trained on LAION 5B, a database of 5.85 billion image caption pairs~\cite{schuhmann2022laion}.  In this paper, we refer to this model as the baseline model.

\paragraph{Datatsets} In order to fine-tune the baseline model, we use the HoliCity dataset~\cite{zhou2020holicity}. This dataset provides 50,078 real images taken in London along with ground truth vanishing points for each image. We use MiDaS~\cite{ranftl2020towards} to compute a depth prediction for each image which is then used as conditioning for the latent diffusion model.\footnote{The HoliCity dataset also provides ground truth depth images, however, they are derived from a CAD model, meaning they are missing finer details such as people, cars, and trees.} This is the same procedure used to originally train the depth-conditioned model~\cite{rombach2022high}. Captions used for conditioning are generated for each image using the BLIP captioning model~\cite{li2022blip}.

\paragraph{Training Details} The code for our fine-tuned model is built using PyTorch on top of~\cite{pinkney2022repo}, which is built on top of the original code released by~\cite{rombach2022high}. The original code from~\cite{pinkney2022repo} is built on top of Stable Diffusion v1, so part of the modifications made by us include updating the code to be compatible with Stable Diffusion v2 checkpoints, including updating the encoder/decoder and dataloaders. We update the loss function of the baseline model to the loss function detailed in Eq.~\ref{eq:new-loss}. We train at an image resolution of 512$\times$512 with a learning rate of 1e-6 and $\lambda = 0.01$. We train for 4 epochs or approximately 200k steps with an effective batch size of 16 after gradient accumulation. We found that the perspective loss had generally saturated by this point. This training takes approximately 12 hours on 4 RTX3090 GPUs. Results are shown in Section~\ref{ssc:latent-diffusion-results}.

\subsubsection{Inpainting}

In addition to text-to-image generation, we also test the value of our constraint for the inpainting task where a model is asked to fill in masked regions of an image. Applying our proposed constraint to the inpainting task does not require any extra training, as we are able to take our general text-to-image diffusion models and perform inpainting using the techniques described by~\cite{lugmayr2022repaint}. We evaluate the results using the LPIPS metric~\cite{zhang2018perceptual} as is the norm for the inpainting task. LPIPS measures the perceptual similarity between two images using features from deep neural networks, in particular AlexNet. Results are shown in Fig.~\ref{fig:inpainting_qual} and Table~\ref{table:inpainting-table} and are discussed in Section~\ref{sec:inpainting_res}

\subsection{Training Monocular Depth Estimation Models}
\label{ssc:depth-est-training}

In order to evaluate the performance from another perspective, we also conduct an experiment on the effect of our new images on monocular depth estimation models. In particular, we fine-tune DPT-Hybrid~\cite{ranftl2021vision} and PixelFormer~\cite{agarwal2023attention} on images generated from both the baseline model and our fine-tuned model. DPT-Hybrid is originally trained on MIX 6, a collection of 10 datasets described in~\cite{ranftl2021vision}, and PixelFormer is originally trained on the KITTI dataset. In order to generate these images, we rely on the SYNTHIA-AL~\cite{bengar2019temporal} and Virtual KITTI 2~\cite{cabon2020vkitti2, gaidon2016virtual} datasets. SYNTHIA-AL contains 70,000 images and Virtual KITTI 2 contains 2,656 images. We take only depth maps from both datasets, and use them as conditioning to generate synthetic images using the base, and our latent diffusion models. In addition, we use BLIP~\cite{li2022blip} to generate captions for all images. For Virtual KITTI 2, we take 8 random crops per image. We also generate diffusion images with 4 different seeds, resulting in a total of 84,992 images derived from the Virtual KITTI 2 dataset. For SYNTHIA, we use the original images, resulting in a total of 70,000 images. Combined, our dataset is 154,992 images and covers various city and driving scenes. We refer to this dataset as All. We additionally train the depth estimation models on images generated only from vKITTI, and refer to this dataset as vKITTI. We additionally append the name of the model used to generate different datasets so that All Enhanced refers to the full set of 155k images generated by our Enhanced model while All Base refers to the full set of images generated by the Baseline model. Results of fine-tuning on these datasets are discussed in Section~\ref{ssc:depth-estimation-results}.

\paragraph{Training Details} For DPT-Hybrid, we train with a learning rate of 5e-6 for 19,500 steps with a batch size of 16. We use a scale and shift invariant loss as described in~\cite{ranftl2021vision,eigen2014depth}. For PixelFormer, we train with a learning rate of 4e-6 for 20,800 steps with a batch size of 8. We train on 1 RTX3090 GPU using the same loss as DPT.

\paragraph{Test Sets} We evaluate the trained depth estimation models on commonly used real datasets KITTI~\cite{geiger2012are} and the outdoor subset of DIODE~\cite{diode_dataset}. We use the Eigen split for KITTI~\cite{eigen2014depth} and a test set of 500 images from DIODE.

\paragraph{Metrics} In order to evaluate the performance of the models, we follow the procedure used by~\cite{ranftl2021vision} and we adopt common depth estimation metrics: Absolute relative error (Abs Rel), Square relative error (Sq Rel), Root mean squared error (RMSE), Log RMSE (RMSE log), and Threshold Accuracy ($\delta_i$) at thresholds $\tau_i$'s = 1.25, $1.25^2$, $1.25^3$ as used in~\cite{agarwal2023attention, ranftl2020towards, ranftl2021vision}.

 \begin{figure}[t]
    \centering
    {\includegraphics[width=.45\textwidth]{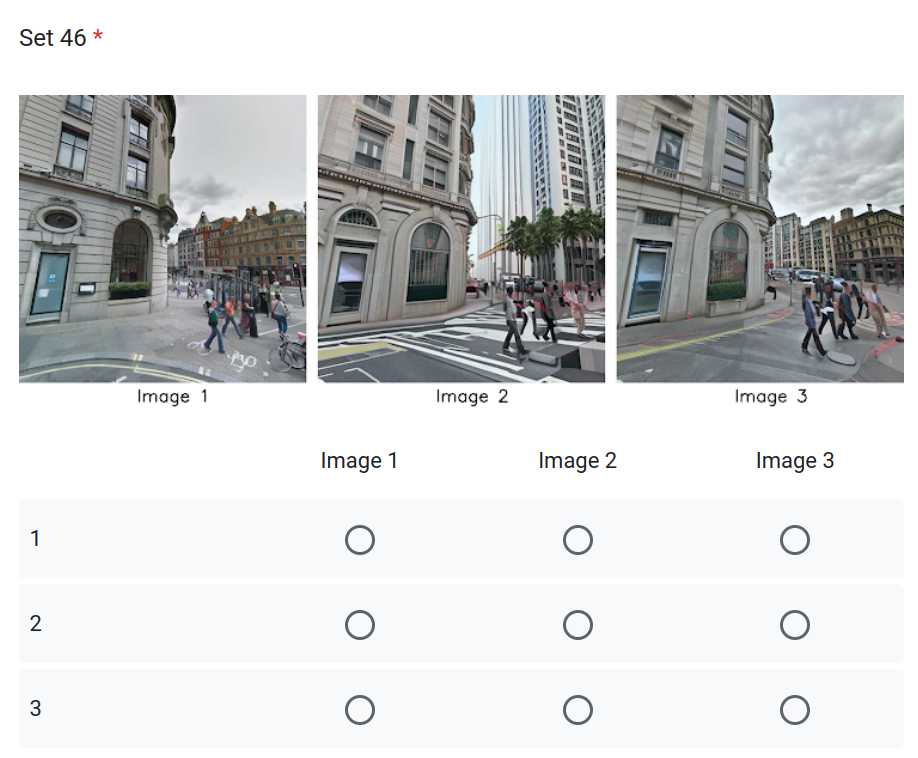}}
    \caption{\textbf{A screenshot of the graphical user interface for the human subjective test we performed on the Prolific platform.} Annotators are asked to rank the image by realism, with "1" being the most and "3" being the least real. Images include one generated from a baseline model, one generated from our enhanced model, and one real image in random order.}
    \label{fig:google_form}
\end{figure}

\subsection{Human Subjective Test Methodology}
\label{ssc:human-subj-methods}

In order to evaluate the photo-realism of images generated by our fine-tuned models, we run human subjective tests on the Prolific~\cite{prolific} website. We ran two tests, one comparing our enhanced model with the baseline model and one comparing our enhanced model with an ablation model. We set up the test as a ranking task where participants are asked to rank sets of three images (Real, Baseline, Ours or Real, Ablation, Ours) in order of photo-realism. The real images come from the HoliCity dataset~\cite{zhou2020holicity}, a landscapes dataset from Kaggle~\cite{landscapedata}, and an animal images dataset from Kaggle~\cite{animalsdata}. The baseline, ablation, and enhanced (ours) images are generated using depth maps extracted from the real image by MiDaS~\cite{ranftl2020towards} and prompts from the BLIP captioning model~\cite{li2022blip}. Participants were shown all three images side by side in random order. Please refer to Fig.~\ref{fig:google_form} for a visualization of the testing setup. We recruit 50 participants across the world and ask them to rate 80 sets of images. Participants were given up to 90 minutes to complete the task. Results from this test are in Section~\ref{ssc:subj-tests-results} and Fig.~\ref{fig:subtest}.

\subsection{Ablation Study}
\label{ssc:ablation-methods}

In order to evaluate the benefits of our proposed constraint, we perform two ablation studies. First, we fine-tune the baseline model on the same dataset but without our updated loss. We refer to this model as the No Loss/Ablation model. We also train a model which takes in vanishing points as conditioning and is trained without our loss. For both models, we generate the same synthetic datasets and train the same monocular depth estimation models described in Section~\ref{ssc:depth-est-training}. Results are shown in Section~\ref{ssc:ablation-results}. An ablation study was also done for the human subjective tests and the inpainting task for the no loss model. Results are described in Section~\ref{ssc:subj-tests-results} and shown in Fig.~\ref{fig:subtest}, Fig.~\ref{fig:inpainting_qual}, and Table~\ref{table:inpainting-table}.

\newcommand{\widthFigure}{.19\textwidth}
\newcommand{\widthPrompt}{.016\textwidth}
\newcommand{\heightSpace}{1pt}
\begin{figure*}
  \centering
  \captionsetup[subfloat]{labelformat=empty, skip=5pt}
  {\includegraphics[width=\widthPrompt]{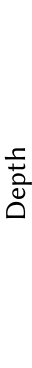}}
  {\includegraphics[width=\widthFigure]{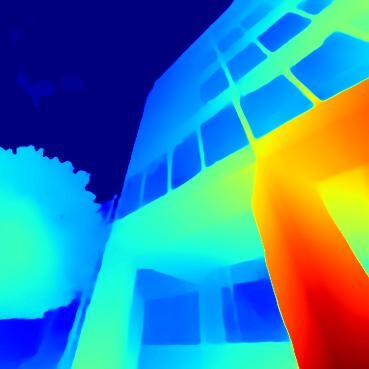}}
  \hfill
  {\includegraphics[width=\widthFigure]{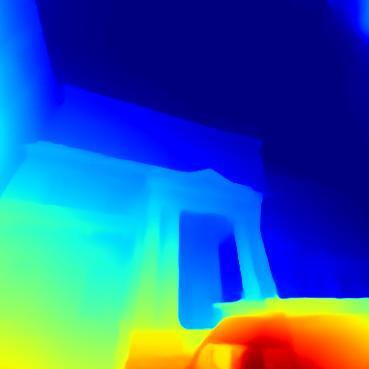}}
  \hfill
  {\includegraphics[width=\widthFigure]{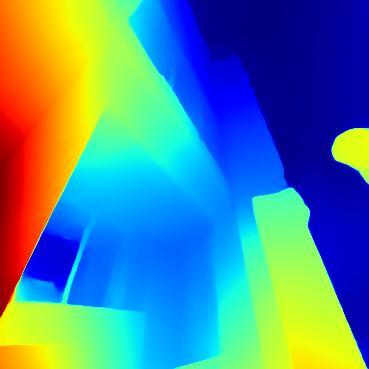}}
  \hfill
  {\includegraphics[width=\widthFigure]{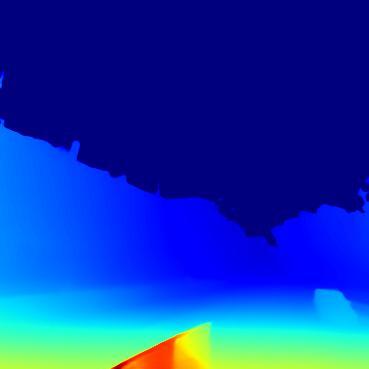}}
  \hfill
  {\includegraphics[width=\widthFigure]{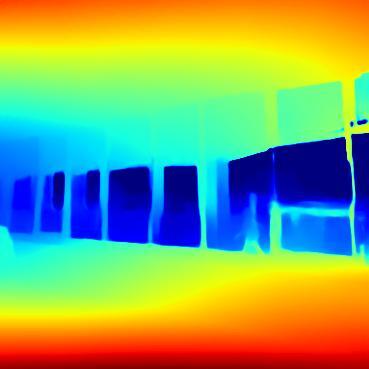}}

  \vspace{\heightSpace}

  {\includegraphics[width=\widthPrompt]{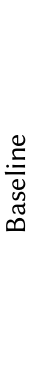}}
  {\includegraphics[width=\widthFigure]{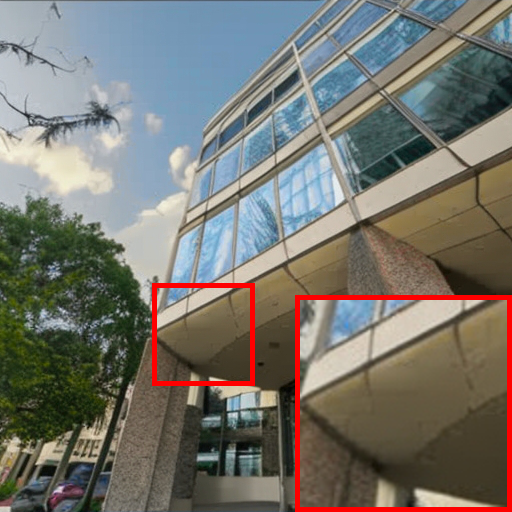}}
  \hfill
  {\includegraphics[width=\widthFigure]{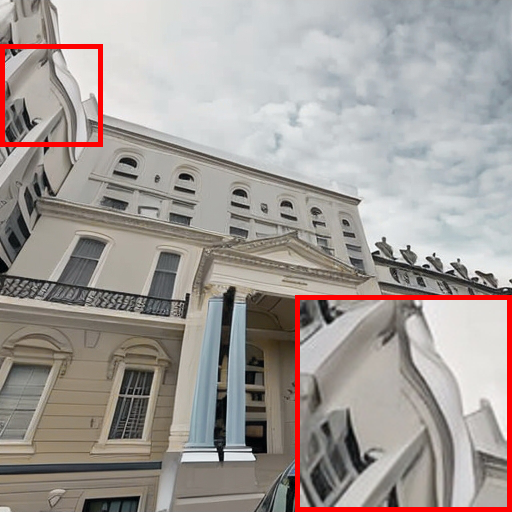}}
  \hfill
  {\includegraphics[width=\widthFigure]{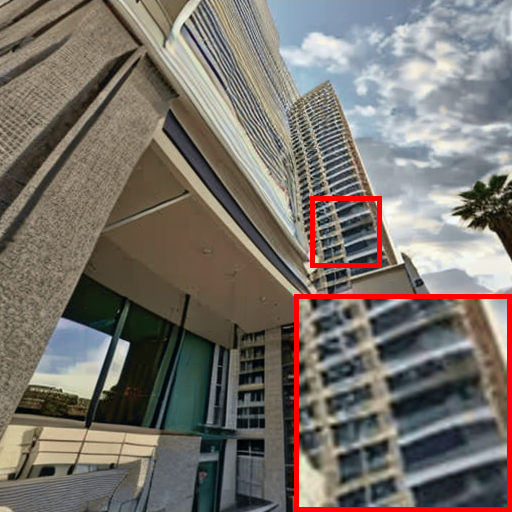}}
  \hfill
  {\includegraphics[width=\widthFigure]{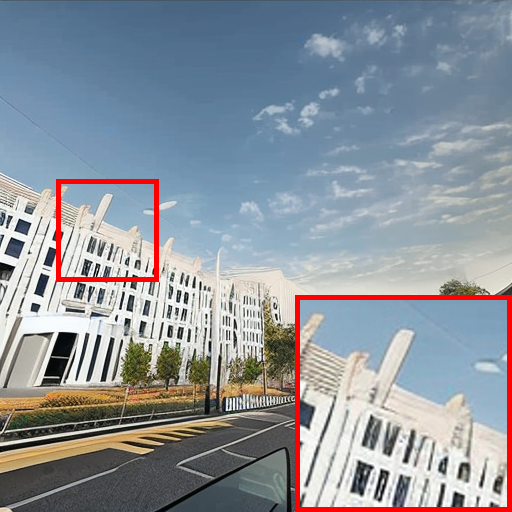}}
  \hfill
  {\includegraphics[width=\widthFigure]{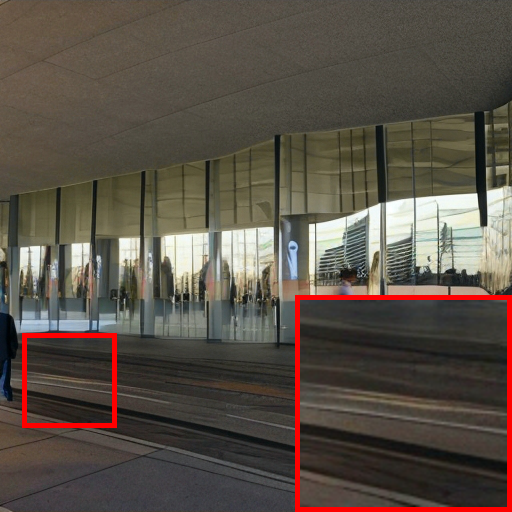}}

  \vspace{\heightSpace}
  
  {\includegraphics[width=\widthPrompt]{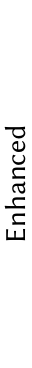}}
  {\includegraphics[width=\widthFigure]{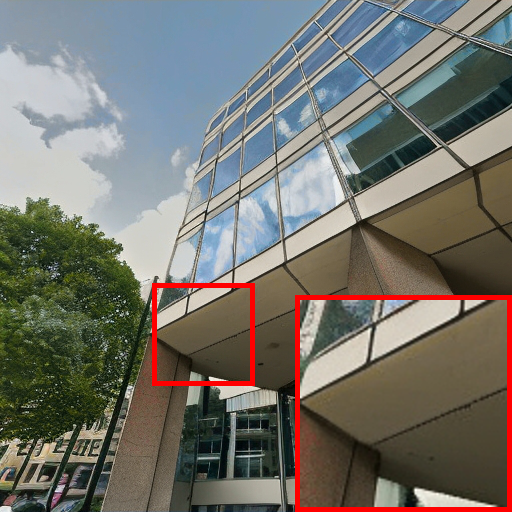}}
  \hfill
  {\includegraphics[width=\widthFigure]{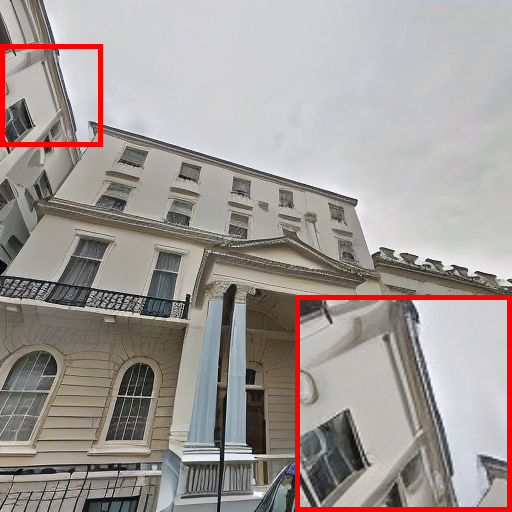}}
  \hfill
  {\includegraphics[width=\widthFigure]{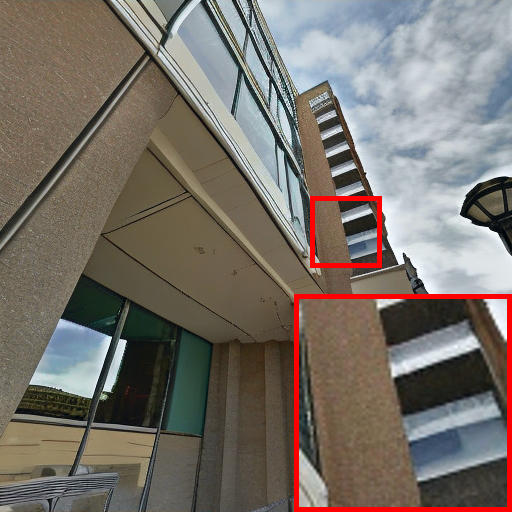}}
  \hfill
  {\includegraphics[width=\widthFigure]{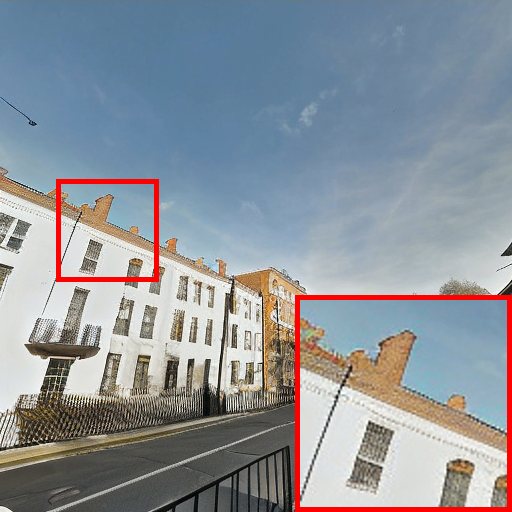}}
  \hfill
  {\includegraphics[width=\widthFigure]{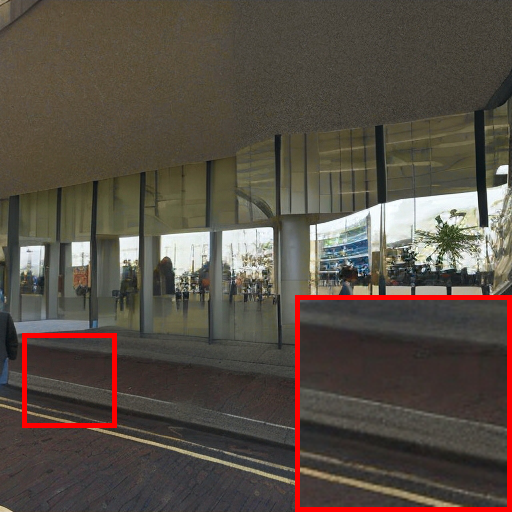}}

  \vspace{\heightSpace}

  {\includegraphics[width=\widthPrompt]{Figures/new_qual_results/depth_prompt.pdf}}
  {\includegraphics[width=\widthFigure]{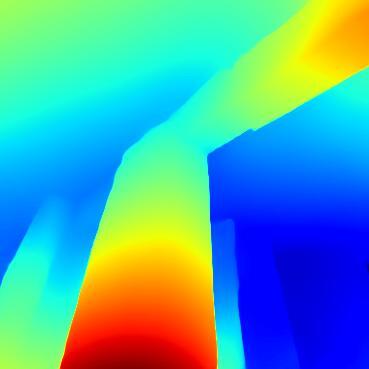}}
  \hfill
  {\includegraphics[width=\widthFigure]{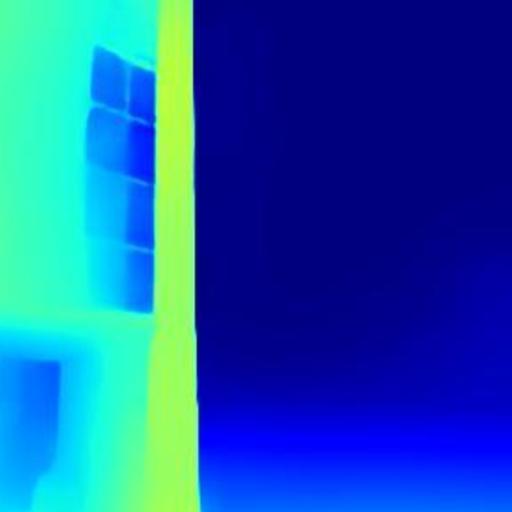}}
  \hfill
  {\includegraphics[width=\widthFigure]{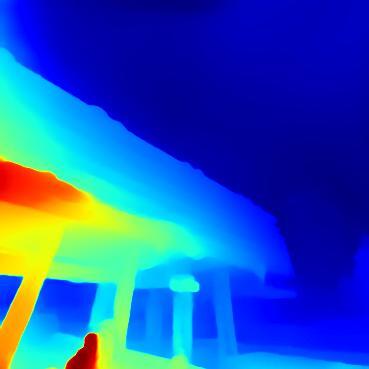}}
  \hfill
  {\includegraphics[width=\widthFigure]{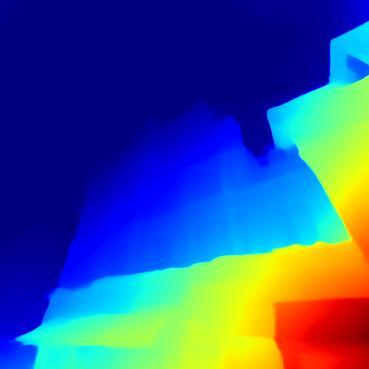}}
  \hfill
  {\includegraphics[width=\widthFigure]{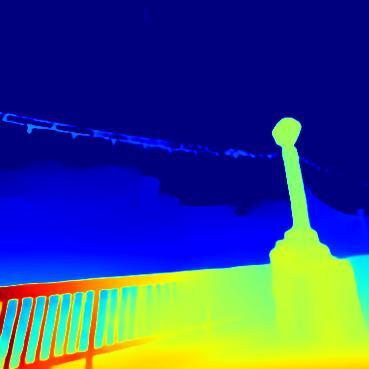}}

  \vspace{\heightSpace}

  {\includegraphics[width=\widthPrompt]{Figures/new_qual_results/baseline_prompt.pdf}}
  {\includegraphics[width=\widthFigure]{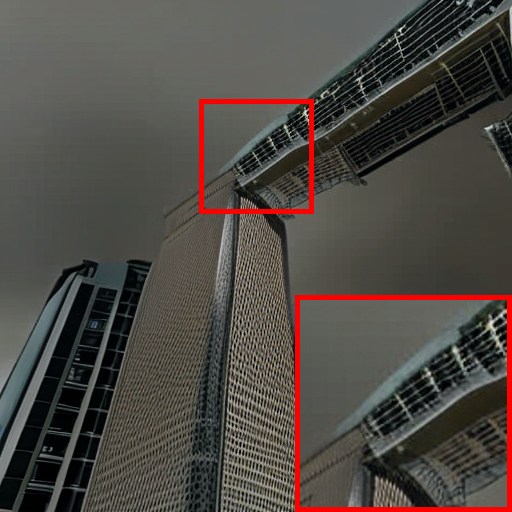}}
  \hfill
  {\includegraphics[width=\widthFigure]{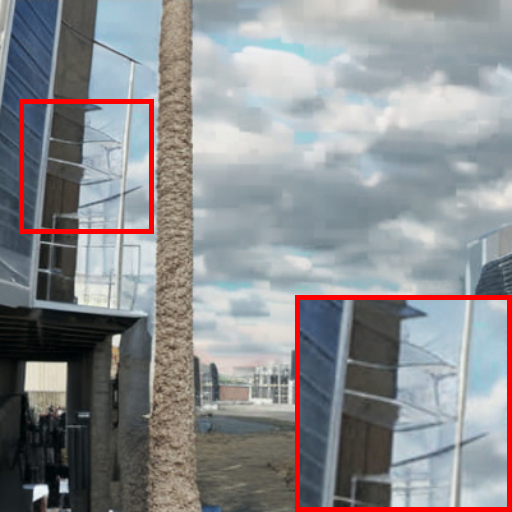}}
  \hfill
  {\includegraphics[width=\widthFigure]{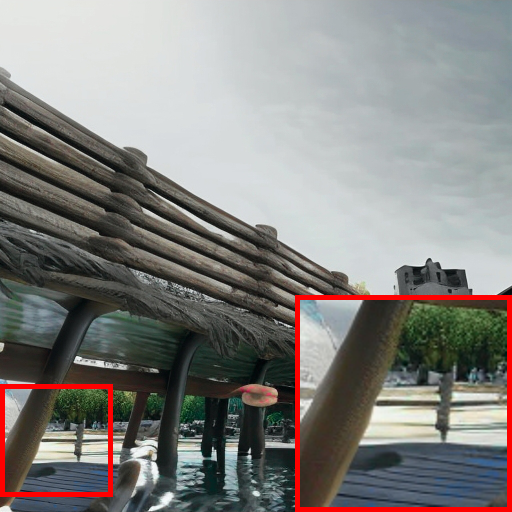}}
  \hfill
  {\includegraphics[width=\widthFigure]{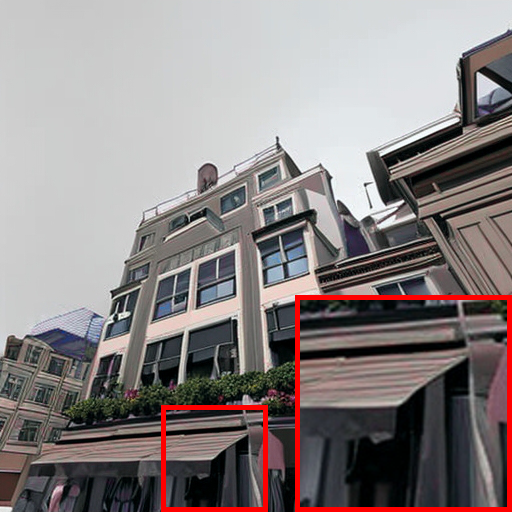}}
  \hfill
  {\includegraphics[width=\widthFigure]{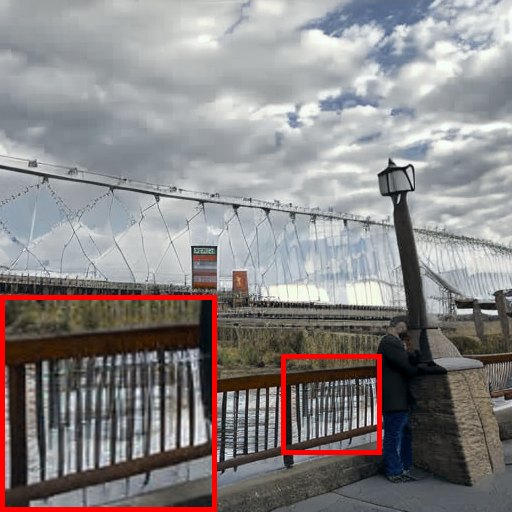}}

  \vspace{\heightSpace}
  
  {\includegraphics[width=\widthPrompt]{Figures/new_qual_results/ours_prompt.pdf}}
  {\includegraphics[width=\widthFigure]{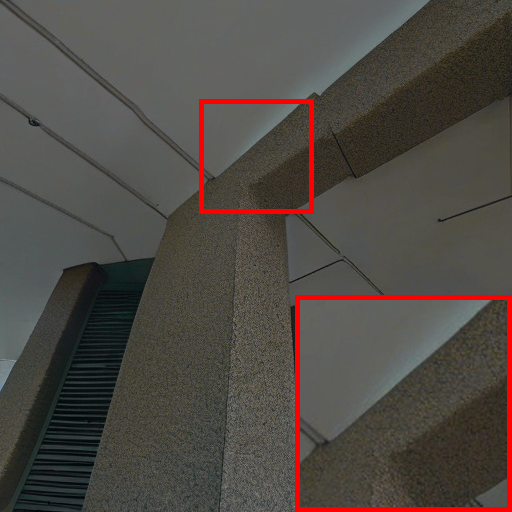}}
  \hfill
  {\includegraphics[width=\widthFigure]{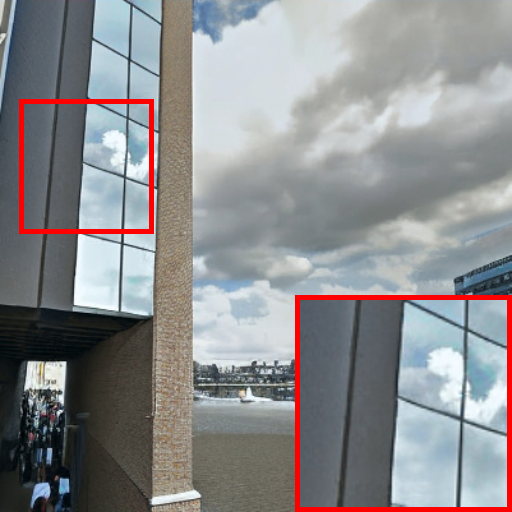}}
  \hfill
  {\includegraphics[width=\widthFigure]{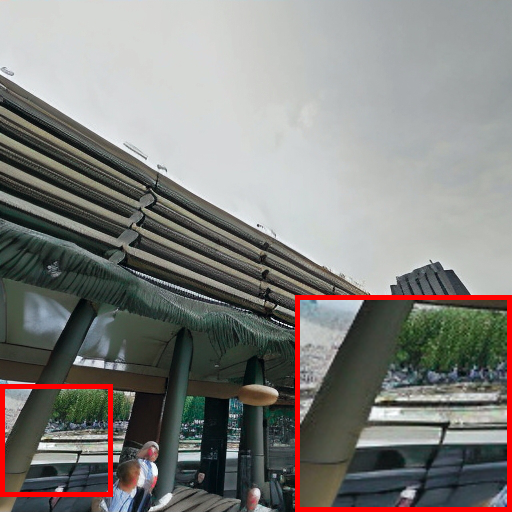}}
  \hfill
  {\includegraphics[width=\widthFigure]{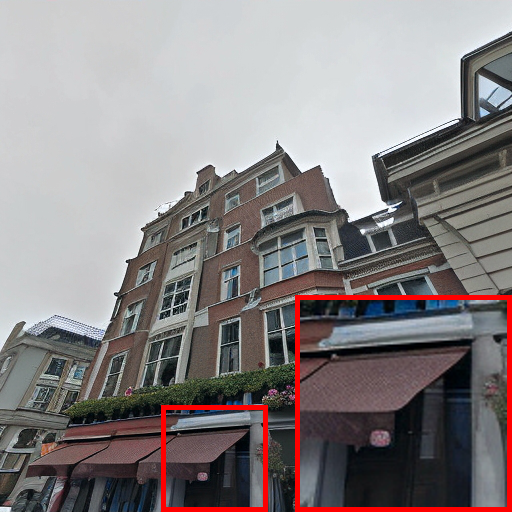}}
  \hfill
  {\includegraphics[width=\widthFigure]{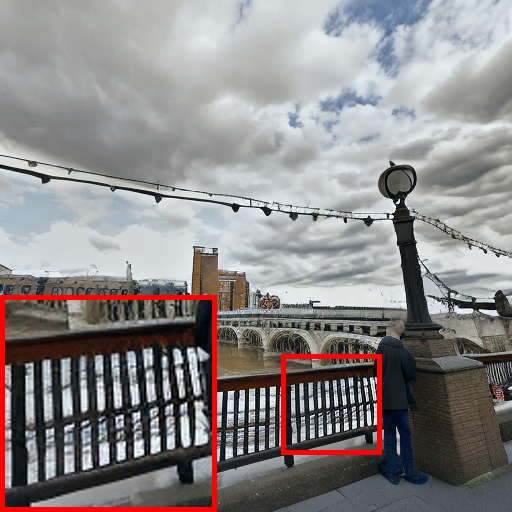}}
  
  \caption{\textbf{Images from our model are better at preserving straight lines.} Examples of outputs from the base model and from our enhanced model. The depth maps these outputs are conditioned on are put at the top. Inlets show specific regions of interest.}
  \label{fig:pespective_qual}
\end{figure*}

\newcommand{\widthGenFigure}{.24\textwidth}
\newcommand{\heightESpace}{3pt}
\begin{figure*}
  \centering
  \captionsetup[subfloat]{labelformat=empty, skip=5pt}
  \vspace{\heightSpace}
  {\includegraphics[width=\widthGenFigure]{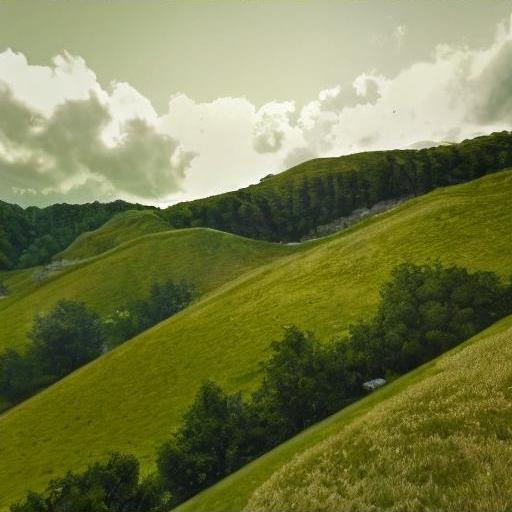}}
  \hfill
  {\includegraphics[width=\widthGenFigure]{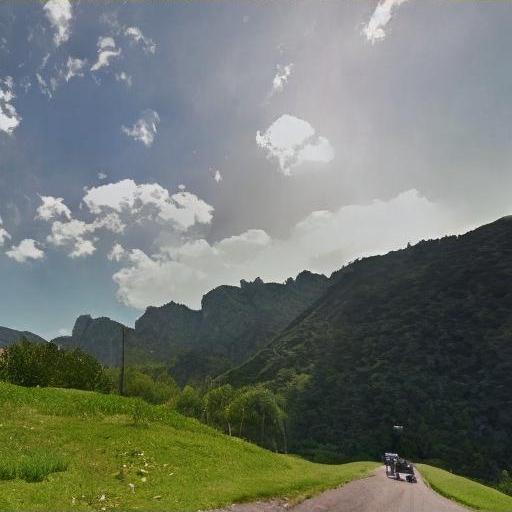}}
  \hfill
  {\includegraphics[width=\widthGenFigure]{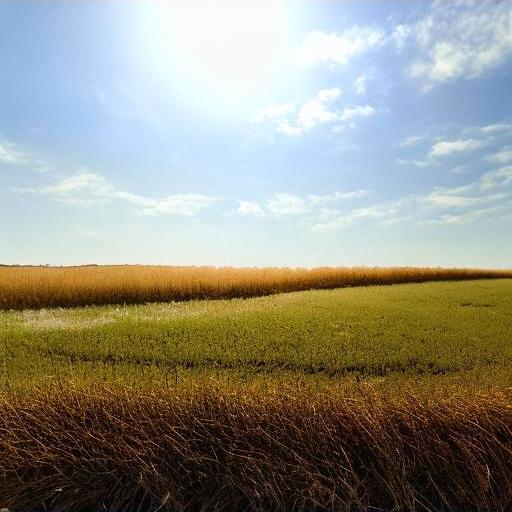}}
  \hfill
  {\includegraphics[width=\widthGenFigure]{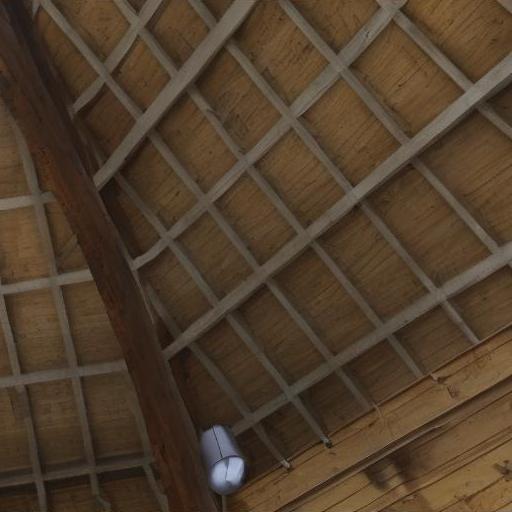}}

  \vspace{\heightESpace}

  {\includegraphics[width=\widthGenFigure]{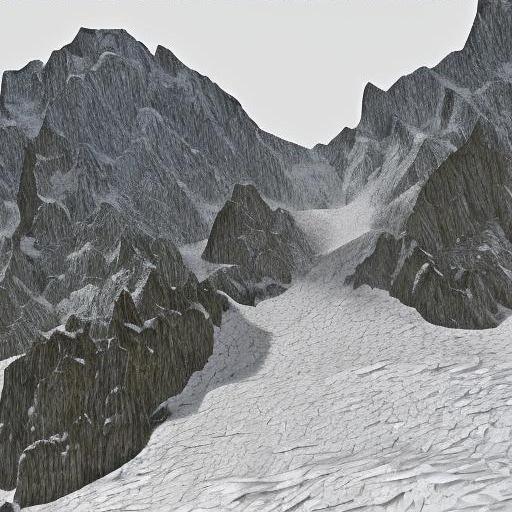}}
  \hfill
  {\includegraphics[width=\widthGenFigure]{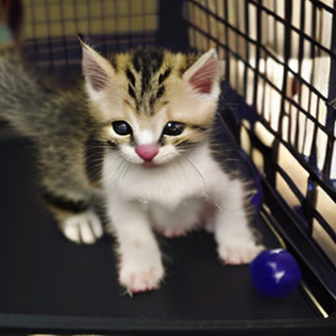}}
  \hfill
  {\includegraphics[width=\widthGenFigure]{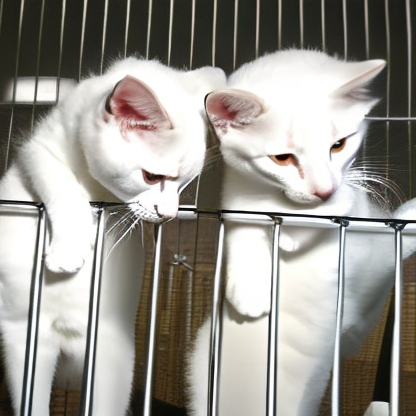}}
  \hfill
  {\includegraphics[width=\widthGenFigure]{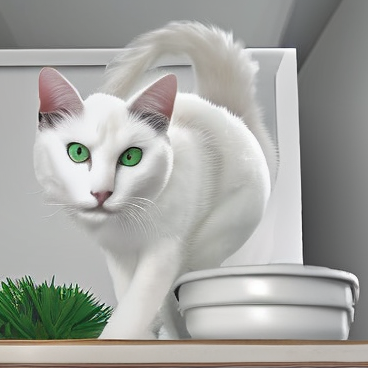}}

  \caption{\textbf{Despite being fine-tuned on images of city scenes, our model is able to generate high-quality images of varied settings including nature landscapes, indoor scenes, and pictures of animals.} Images were taken from a landscapes dataset~\cite{landscapedata}, an animal dataset~\cite{animalsdata}, and the indoor subset of the DIODE dataset~\cite{diode_dataset}. }
  \label{fig:pespective_qual_gen}
\end{figure*}

\begin{figure*}
  \centering
  \captionsetup[subfloat]{labelformat=empty, skip=5pt}
  {\includegraphics[width=\widthFigure]{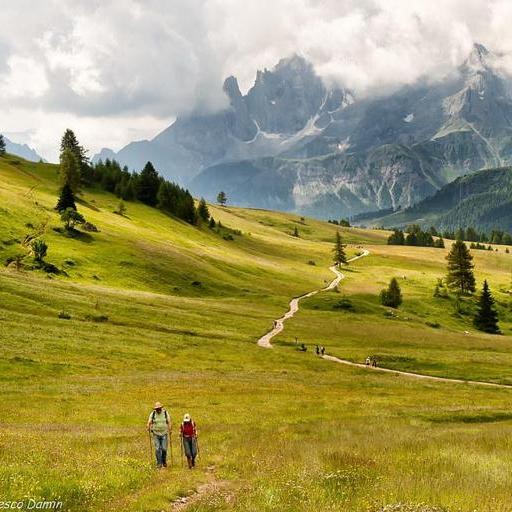}}
  \hfill
  {\includegraphics[width=\widthFigure]{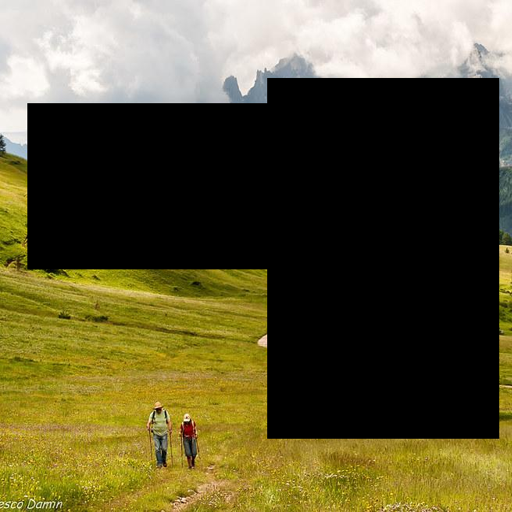}}
  \hfill
  {\includegraphics[width=\widthFigure]{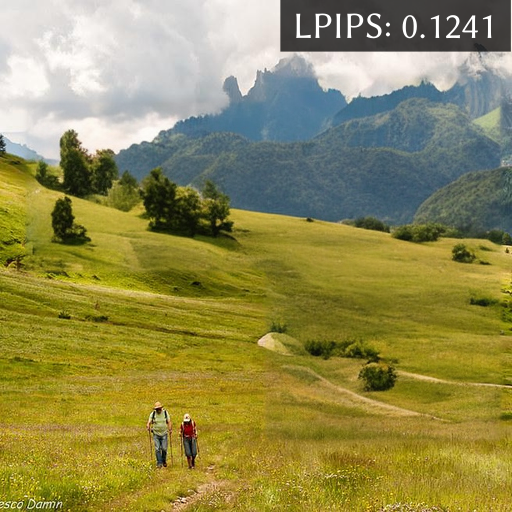}}
  \hfill
  {\includegraphics[width=\widthFigure]{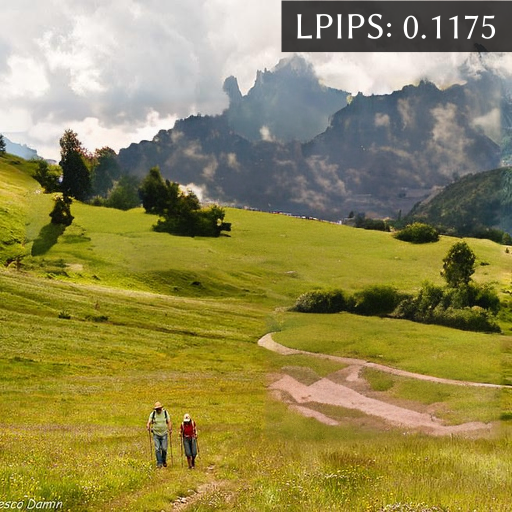}}
  \hfill
  {\includegraphics[width=\widthFigure]{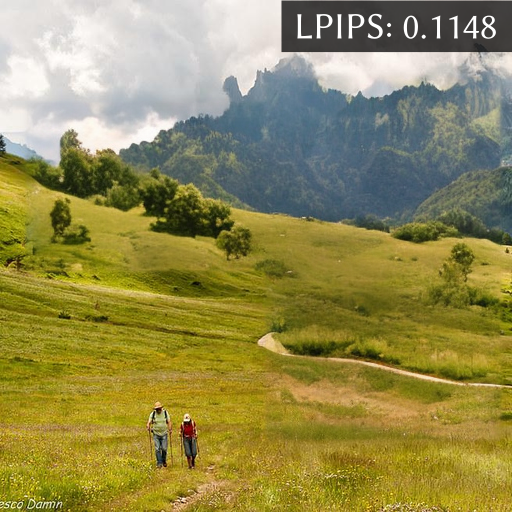}}

  \vspace{\heightESpace}
  
  \subfloat[\centering Original]
  {\includegraphics[width=\widthFigure]{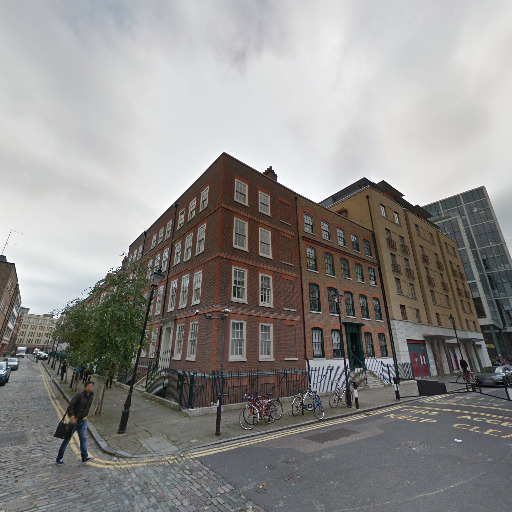}}
  \hfill
  \subfloat[\centering Masked]
  {\includegraphics[width=\widthFigure]{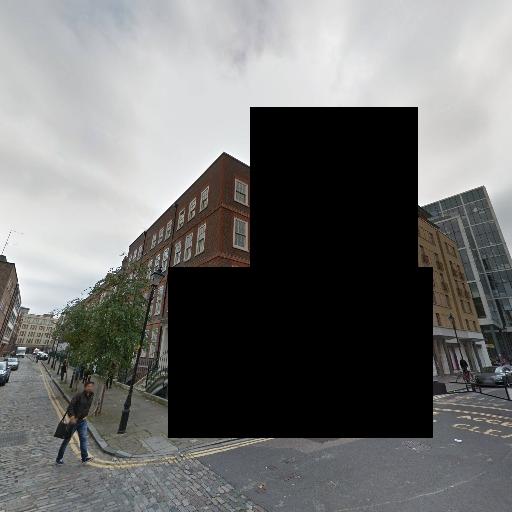}}
  \hfill
  \subfloat[\centering Baseline]
  {\includegraphics[width=\widthFigure]{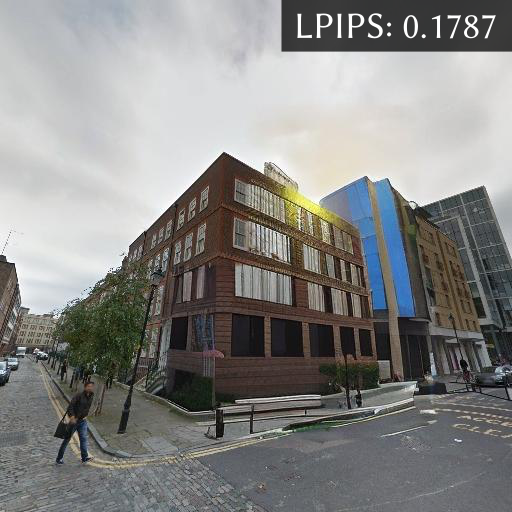}}
  \hfill
  \subfloat[\centering No Loss (Ablation)]
  {\includegraphics[width=\widthFigure]{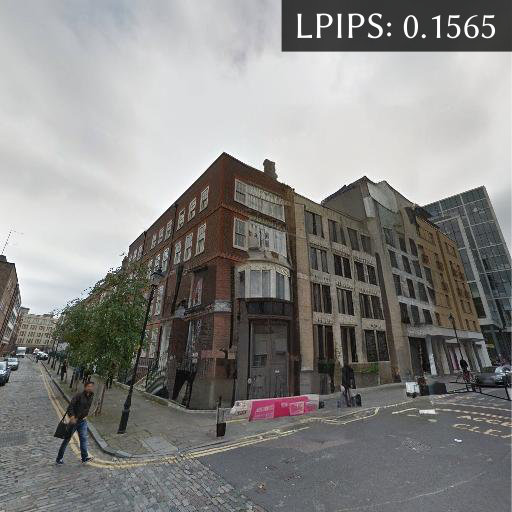}}
  \hfill
  \subfloat[\centering Enhanced (Ours)] 
  {\includegraphics[width=\widthFigure]{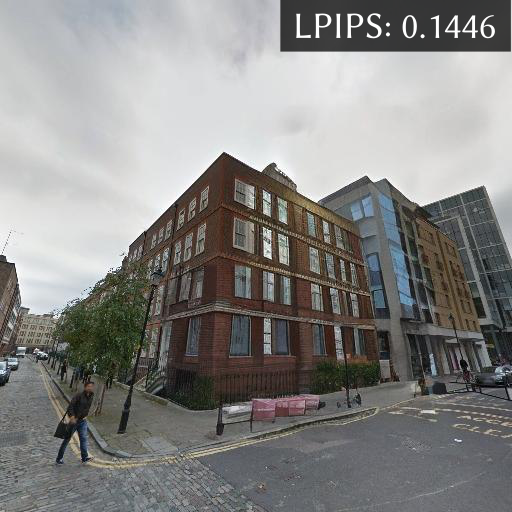}}

  \caption{\textbf{The proposed geometric constraint provides benefits for the inpainting task on diverse scenes.} Images reconstructed with our enhanced model consistently outperform the baseline and ablation models on LPIPS scores (shown in the top right, lower is better). }
  \label{fig:inpainting_qual}
\end{figure*}

\section{Results}

This results section is split into sub-sections according to the experiments described in Section 5. In Section~\ref{ssc:latent-diffusion-results}, we describe the results of fine-tuning latent diffusion models.  In Section~\ref{ssc:depth-estimation-results}, we discuss the results of fine-tuning SOTA monocular depth estimation models on our generated images. In Section~\ref{ssc:subj-tests-results}, we discuss the results of our human subjective test, and in Section~\ref{ssc:ablation-results}, we discuss the results of our ablation tests.

\newcommand{\widthFigureAblation}{.163\textwidth}
\begin{figure*}
  \centering
  \captionsetup[subfloat]{labelformat=empty, skip=1pt}
  \subfloat[\centering Depth]{\includegraphics[width=\widthFigureAblation]{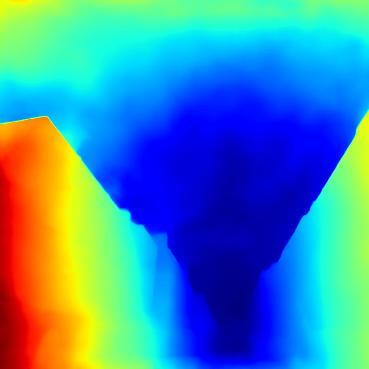}}
  \hfill
  \subfloat[\centering Baseline]{\includegraphics[width=\widthFigureAblation]{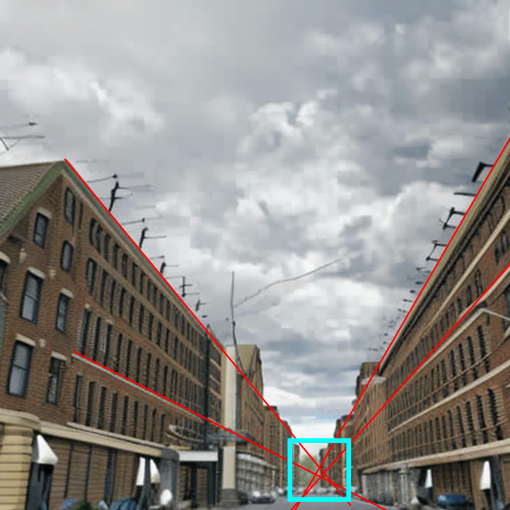}}
  \hfill
  \subfloat[\centering Enhanced]{\includegraphics[width=\widthFigureAblation]{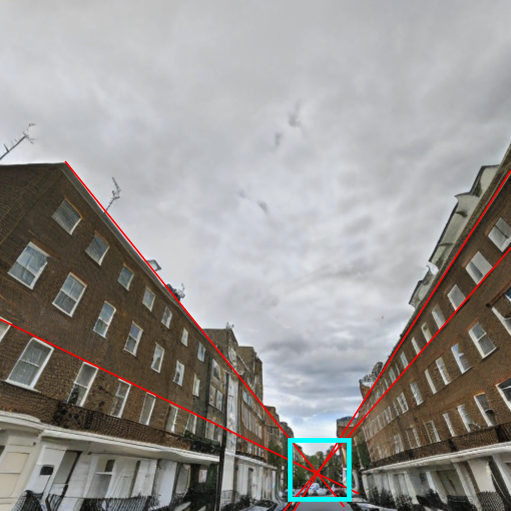}}
  \hspace{3pt}
  \subfloat[\centering Depth]{\includegraphics[width=\widthFigureAblation]{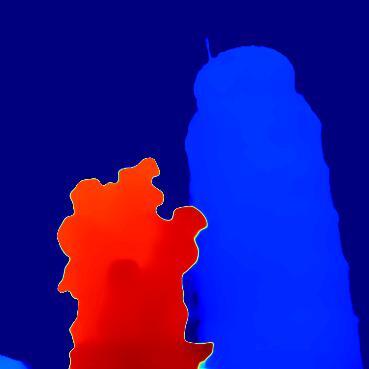}}
  \hfill
  \subfloat[\centering Baseline]{\includegraphics[width=\widthFigureAblation]{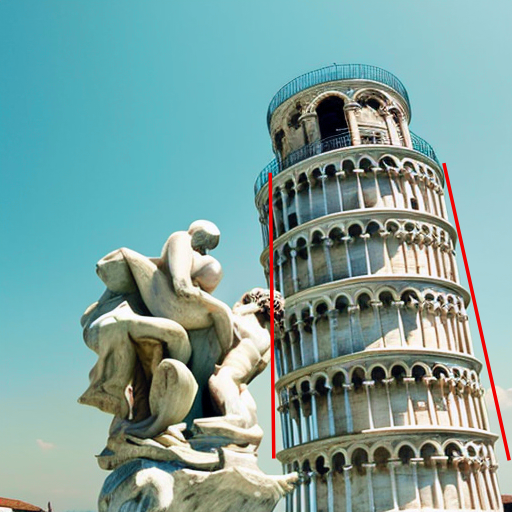}}
  \hfill
  \subfloat[\centering Enhanced]{\includegraphics[width=\widthFigureAblation]{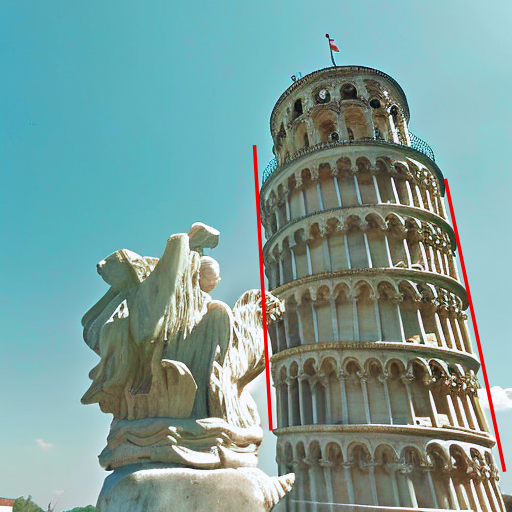}}
  
  \caption{\textbf{Images from our model have more consistent vanishing point lines.} This figure shows examples of stable diffusion outputs from the baseline model and from our model with perspective loss along with perspective lines for the image. The depth maps these outputs are conditioned on are put in the left-hand column. Note that for the baseline image in the first row, the lines do not intersect at a single vanishing point, violating perspective geometry. These violations can sometimes result in curved lines as seen in the baseline image in the second row.}
  \label{fig:perspective_qual_lines}
\end{figure*}

\subsection{Fine-tuned Latent Diffusion Models}
\label{ssc:latent-diffusion-results}

We show some representative generations from our fine-tuned model in Fig.~\ref{fig:pespective_qual}. In the figure, we show the depth maps used to condition the diffusion models along with generations from the baseline model and our enhanced model. Images from the baseline model tend to suffer from curved lines and distortions that affect perspective accuracy. In particular, the baseline model tends to have trouble accurately generating regions with windows, high-frequency details such as many parallel horizontal or vertical lines, and corners. We also draw perspective lines on images from the baseline and our models in Fig.~\ref{fig:perspective_qual_lines}. Images from our model tend to have more coherent perspective lines and more accurate vanishing points. In addition, in both figures, because of the aforementioned distortions, the baseline images look further from the distribution of natural images than images from our model. Since our enhanced model is fine-tuned on a dataset of mainly only cityscapes, we also generate varied nature~\cite{landscapedata}, animal~\cite{animalsdata}, and indoor scenes~\cite{diode_dataset} to verify that this fine-tuning does not limit the ability of the model to generate other types of images. Some representative images are shown in Fig.~\ref{fig:pespective_qual_gen}. We additionally quantitatively evaluate these images using the FID metric~\cite{heusel2017gans}. Our model outperforms both the baseline model and the no loss model. The results are shown in Table~\ref{table:FID-table}.

\subsubsection{Inpainting}
\label{sec:inpainting_res}

We evaluate the inpainting performance of our models using both qualitative (Fig.~\ref{fig:inpainting_qual}) and quantitative (Table~\ref{table:inpainting-table}) results. All three models of interest, the baseline model, ablation model, and enhanced model were tested on the combination of two datasets: the HoliCity validation set~\cite{zhou2020holicity} and a landscape dataset~\cite{landscapedata}. The LPIPS metric~\cite{zhang2018perceptual}, which measures perceptual similarity using features from deep image networks, was used to compare models as is the norm for the inpainting task. We used the official implementation provided by~\cite{zhang2018perceptual}. Note that lower is better for the LPIPS metric. As seen in Table~\ref{table:inpainting-table}, our enhanced model consistently outperforms both the baseline model and ablation model, with a 7.1\% improvement over the baseline model and a 3.6\% improvement over the ablation model on the combined dataset.

\newcommand{\widthdepthfigure}{.33\textwidth}
\newcommand{\heightdepthSpace}{4pt}
\begin{figure*}
  \centering
    \captionsetup[subfloat]{labelformat=empty, skip=0pt}
  {\includegraphics[width=\widthdepthfigure]{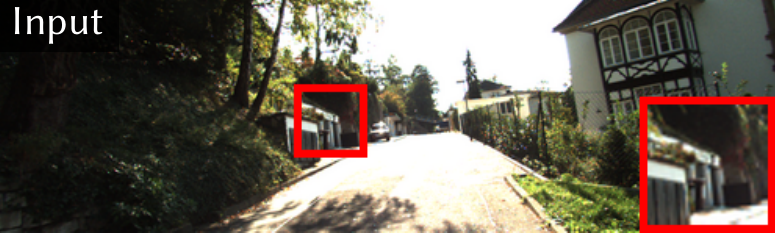}}
  \hfill
  {\includegraphics[width=\widthdepthfigure]{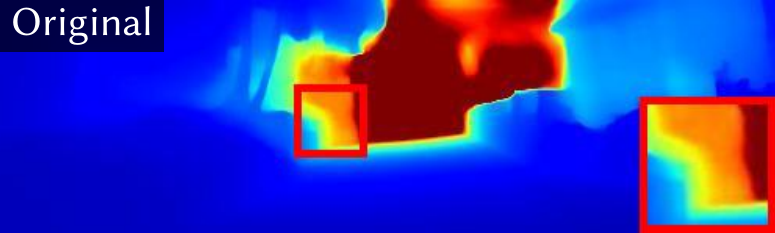}}
  \hfill
  {\includegraphics[width=\widthdepthfigure]{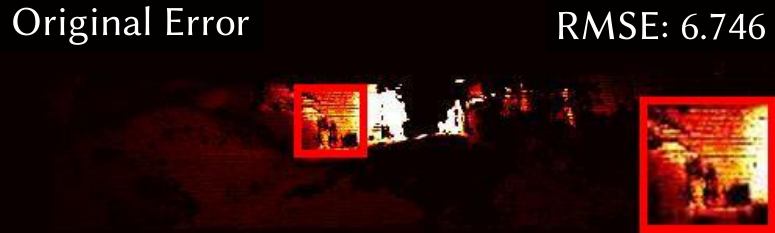}}

  \vspace{\heightSpace}

  {\includegraphics[width=\widthdepthfigure]{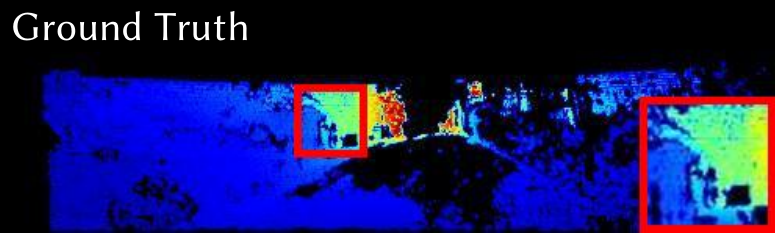}}
  \hfill
  {\includegraphics[width=\widthdepthfigure]{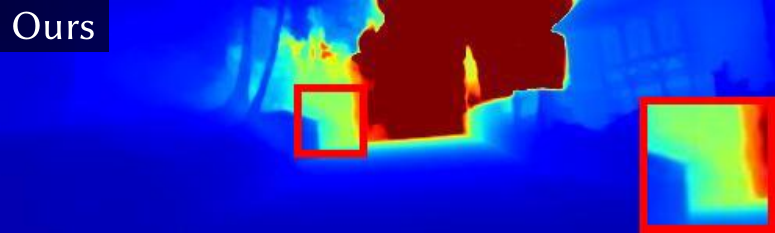}}
  \hfill
  {\includegraphics[width=\widthdepthfigure]{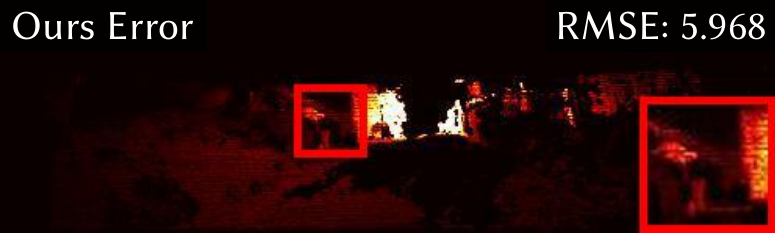}}

  \vspace{\heightdepthSpace}

  {\includegraphics[width=\widthdepthfigure]{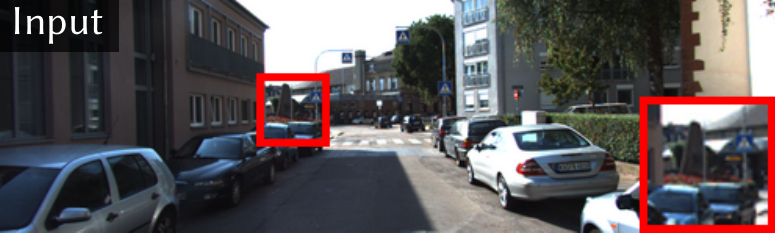}}
  \hfill
  {\includegraphics[width=\widthdepthfigure]{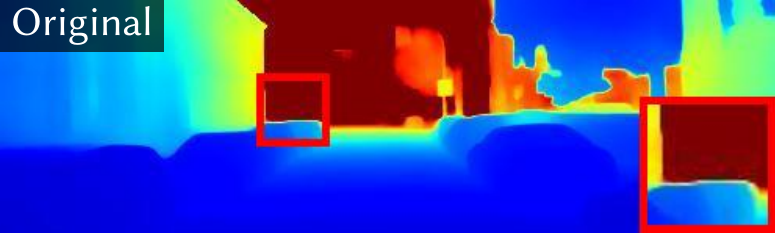}}
  \hfill
  {\includegraphics[width=\widthdepthfigure]{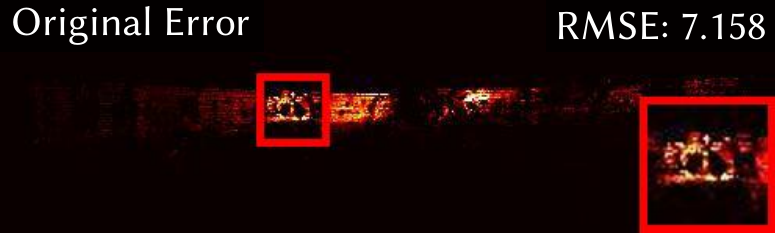}}

  \vspace{\heightSpace}

  {\includegraphics[width=\widthdepthfigure]{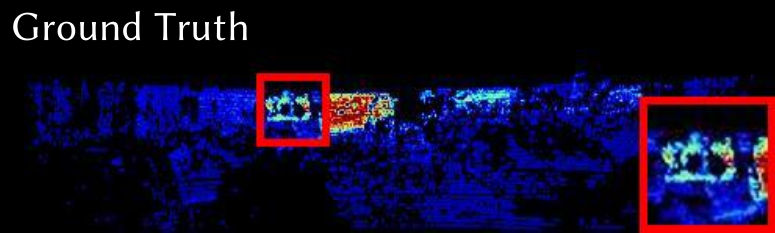}}
  \hfill
  {\includegraphics[width=\widthdepthfigure]{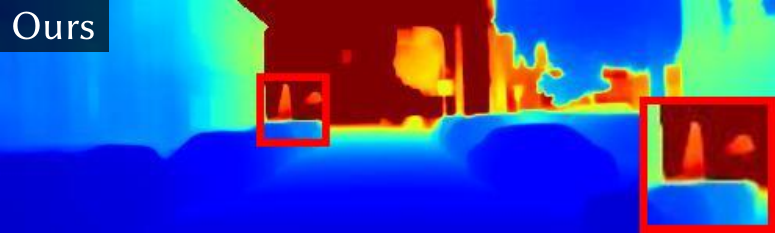}}
  \hfill
  {\includegraphics[width=\widthdepthfigure]{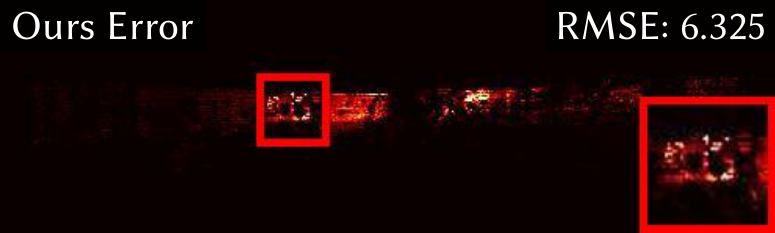}}

  \vspace{\heightdepthSpace}

  {\includegraphics[width=\widthdepthfigure]{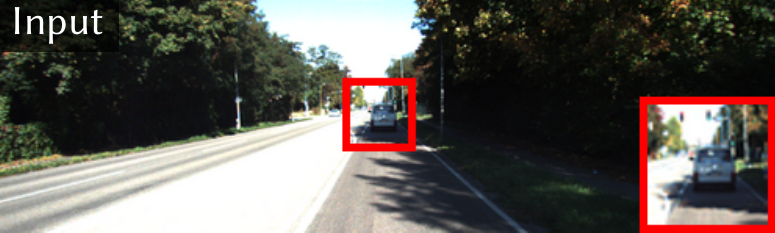}}
  \hfill
  {\includegraphics[width=\widthdepthfigure]{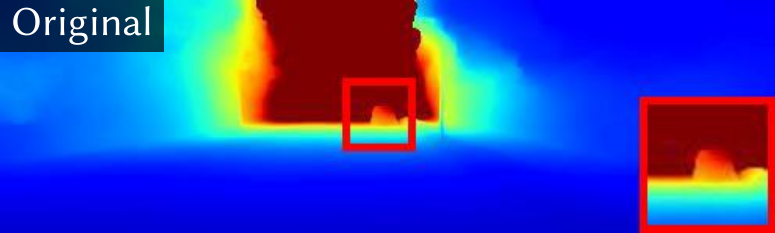}}
  \hfill
  {\includegraphics[width=\widthdepthfigure]{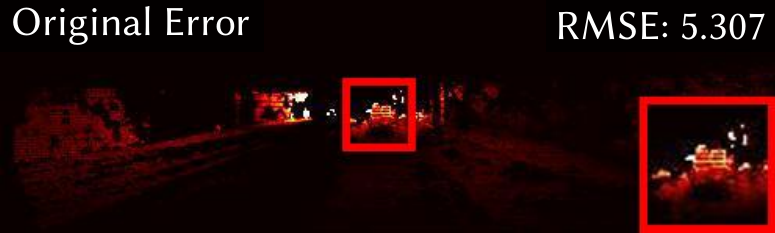}}

  \vspace{\heightSpace}

  {\includegraphics[width=\widthdepthfigure]{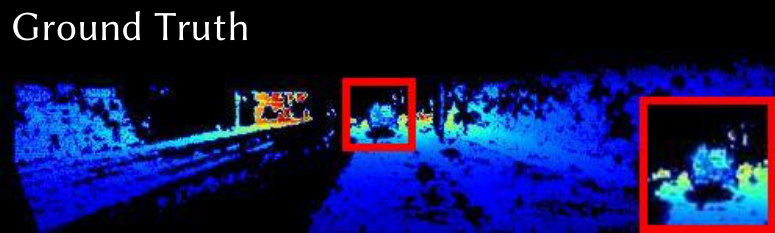}}
  \hfill
  {\includegraphics[width=\widthdepthfigure]{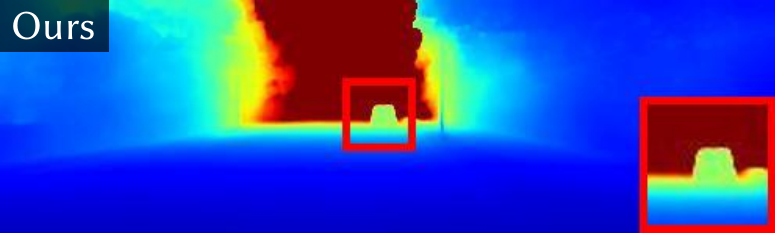}}
  \hfill
  {\includegraphics[width=\widthdepthfigure]{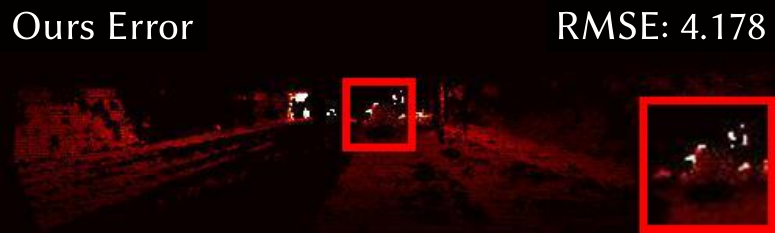}}

   \vspace{\heightdepthSpace}

  {\includegraphics[width=\widthdepthfigure]{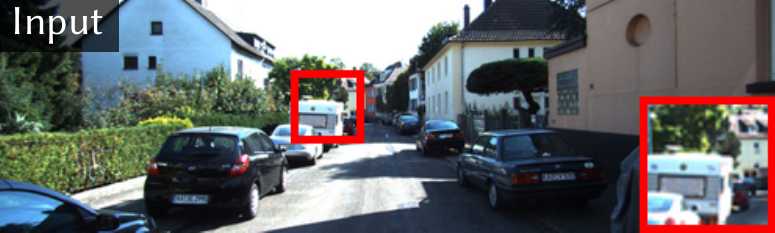}}
  \hfill
  {\includegraphics[width=\widthdepthfigure]{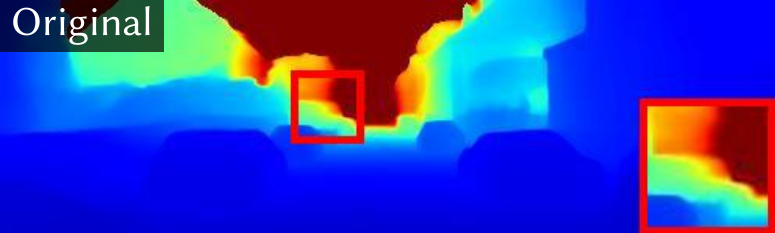}}
  \hfill
  {\includegraphics[width=\widthdepthfigure]{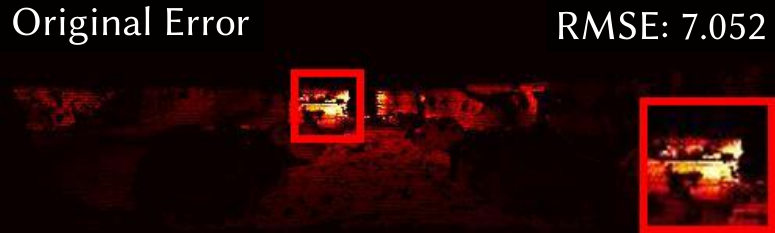}}

  \vspace{\heightSpace}

  {\includegraphics[width=\widthdepthfigure]{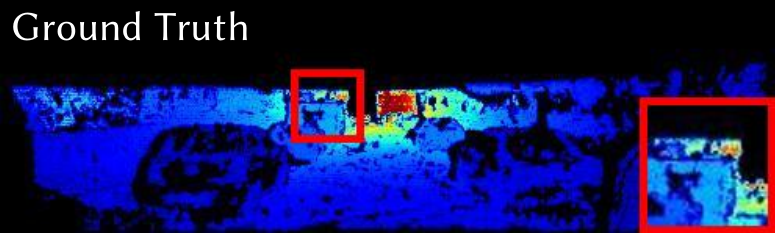}}
  \hfill
  {\includegraphics[width=\widthdepthfigure]{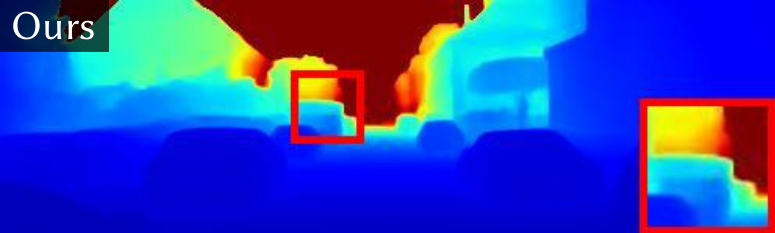}}
  \hfill
  {\includegraphics[width=\widthdepthfigure]{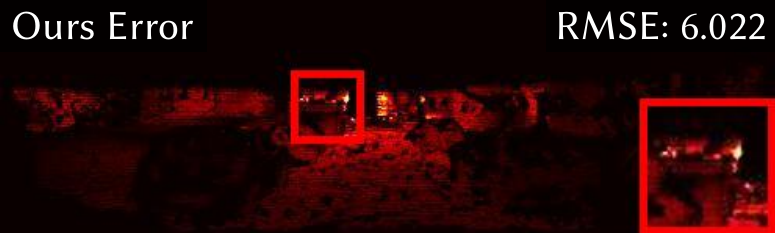}}

  \caption{\textbf{Qualitative comparisons of DPT-Hybrid fine-tuned on the data from our fine-tuned models and the original DPT-Hybrid model.} The depth maps produced by models trained on images from our enhanced model capture more high-frequency detail than the models trained on images from the baseline model. The RMSE error of the outputs of our model is also consistently lower.}
  \label{fig:depth_qual}
\end{figure*}

\subsection{Monocular Depth Estimation}
\label{ssc:depth-estimation-results}

\begin{figure*}
  \centering
  \captionsetup[subfloat]{labelformat=empty, skip=1pt}
  \subfloat[\centering Depth]{\includegraphics[width=\widthFigureAblation]{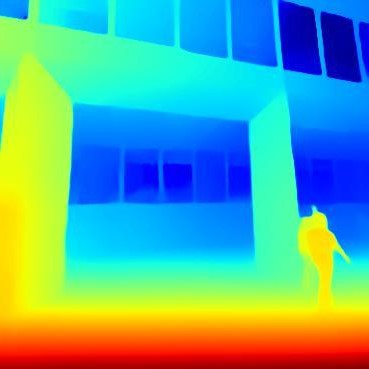}}
  \hfill
  \subfloat[\centering No Loss]{\includegraphics[width=\widthFigureAblation]{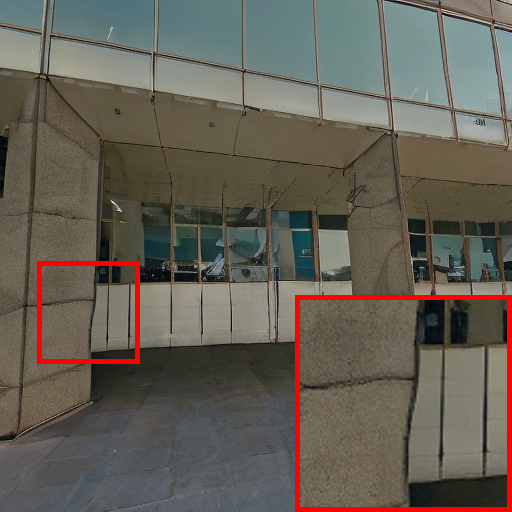}}
  \hfill
  \subfloat[\centering Enhanced]{\includegraphics[width=\widthFigureAblation]{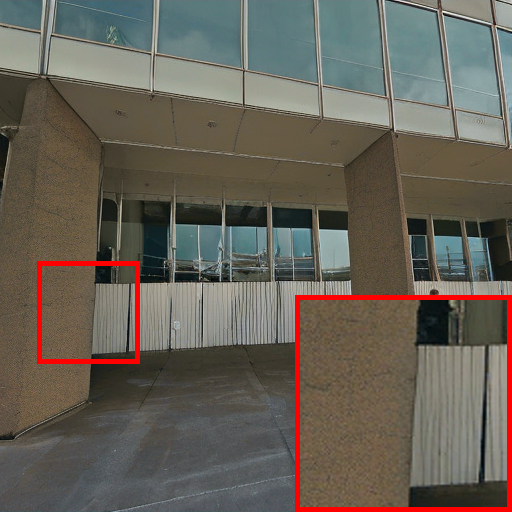}}
  \hspace{3pt}
  \subfloat[\centering Depth]{\includegraphics[width=\widthFigureAblation]{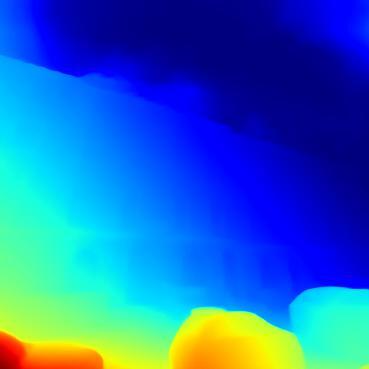}}
  \hfill
  \subfloat[\centering No Loss]{\includegraphics[width=\widthFigureAblation]{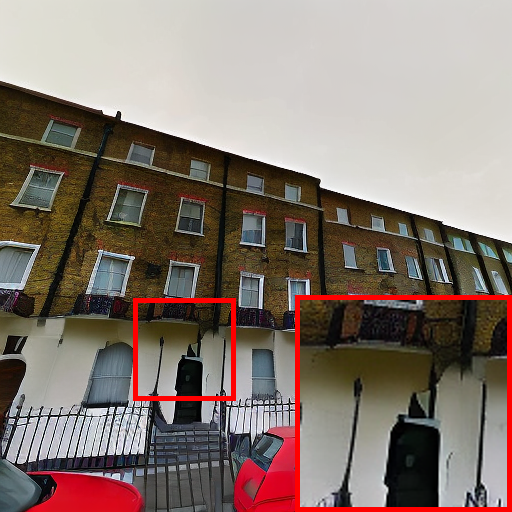}}
  \hfill
  \subfloat[\centering Enhanced]{\includegraphics[width=\widthFigureAblation]{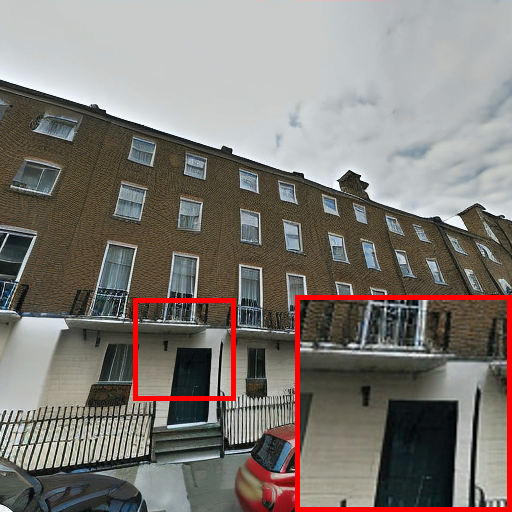}}
  
  \caption{\textbf{The proposed perspective constraint is responsible for the increase in perspective accuracy of generated images more than the dataset the diffusion models were fine-tuned on.} The depth maps these outputs are conditioned on are put in the left-hand column. Note that the images without our loss suffer from more distortions and curved lines and are less photo-realistic. }
  \label{fig:pespective_qual_ablation}
\end{figure*}

\begin{table*}
\vspace{10pt}
  \caption{\textbf{Monocular Depth Estimation performance of DPT-Hybrid fine-tuned on our data compared to the base DPT-Hybrid model.} The original DPT-Hybrid model was trained on a dataset referred to as MIX 6, which is a collection of 10 datasets as described in~\cite{ranftl2021vision}. Fine-tuned models were trained on synthetic datasets generated by either the base stable diffusion model or our fine-tuned model. The best performing model is in bold and the second best is underlined.}
  \label{tab:DPT-quant}
  \small
  \begin{tabular}{c|c|c|cccccccc}
    \toprule
    Model & Description & Test Set & RMSE $\downarrow$ & RMSE log $\downarrow$ & AbsRel $\downarrow$ & SqRel $\downarrow$ & SiLog $\downarrow$ & $\delta_1$ $\uparrow$ & $\delta_2$ $\uparrow$ & $\delta_3$ $\uparrow$ \\
    \midrule
    \multirow{3}{*}{DPT-Hybrid} & Original & \multirow{3}{*}{KITTI} & 5.0287 & 0.1874 & 0.1328 & 0.9705 & 18.6320 & 0.8385 & 0.9552 & 0.9855 \\ 
    & Fine-tuned on vKITTI Base & & \underline{4.7680} & \underline{0.1800} & \underline{0.1286} & \underline{0.8104} & \underline{17.8890} & \underline{0.8401} & \underline{0.9587} & \underline{0.9881}\\
    & Fine-tuned on vKITTI Enhanced &  &  \textbf{4.6749} & \textbf{0.1760} & \textbf{0.1250} & \textbf{0.7827} & \textbf{17.4836} & \textbf{0.8496} & \textbf{0.9608} & \textbf{0.9890}\\
    \midrule
    \multirow{3}{*}{DPT-Hybrid} & Original & \multirow{3}{*}{\begin{tabular}{c}DIODE \\ Outdoor\end{tabular}} & 9.5311 & \underline{0.5667} & 0.4593 & 7.0644 & \underline{52.6255} & \underline{0.4709} & \underline{0.6588} & \underline{0.7759} \\
    & Fine-tuned on vKITTI Base &  & \underline{9.4863} & 0.5669 & \underline{0.4560} & \textbf{6.7930} & 52.6316 & 0.4705 & 0.6586 & 0.7758\\
    & Fine-tuned on vKITTI Enhanced & & \textbf{9.4854} & \textbf{0.5663} & \textbf{0.4559} & \underline{6.8371} & \textbf{52.5902} & \textbf{0.4713} & \textbf{0.6595} & \textbf{0.7763} \\
    \bottomrule
  \end{tabular}
  \label{table:dpt-kitti-results}
\end{table*}

\begin{table*}
    \vspace{10pt}
  \caption{\textbf{Monocular Depth Estimation performance of PixelFormer fine-tuned on our data compared to the base PixelFormer model (trained on KITTI) on the DIODE outdoor dataset.} Fine-tuned models were trained on synthetic datasets generated by either the base stable diffusion model or our fine-tuned model. The best performing model is in bold and the second best is underlined.}
  \label{tab:pixelformer-quant}
  \small
  \begin{tabular}{c|c|c|cccccccc}
    \toprule
    Model & Description & Test Set & RMSE $\downarrow$ & RMSE log $\downarrow$ & AbsRel $\downarrow$ & SqRel $\downarrow$ & SiLog $\downarrow$ & $\delta_1$ $\uparrow$ & $\delta_2$ $\uparrow$ & $\delta_3$ $\uparrow$ \\
    \midrule
    \multirow{4}{*}{PixelFormer} & Original & \multirow{4}{*}{\begin{tabular}{c}DIODE \\ Outdoor\end{tabular}} & 8.8726 & 0.7041 & 1.4532 & 21.8911 & 66.0165 & 0.3254 & 0.5586 & 0.7075 \\
    & Fine-tuned on KITTI & & 8.9302 & 0.7102 & 1.4441 & 22.1350 & 66.4702 & 0.3244 & 0.5523 & 0.6929 \\
    & Fine-tuned on vKITTI Base & & \underline{8.5381} & \underline{0.6891} & \underline{1.4140} & \underline{21.8363} & \underline{64.5891} & \underline{0.3294} & \underline{0.5651} & \underline{0.7209} \\ 
    & Fine-tuned on vKITTI Enhanced & & \textbf{8.4728} & \textbf{0.6870} & \textbf{1.3738} & \textbf{19.3406} & \textbf{64.4721} & \textbf{0.3329} & \textbf{0.5677} & \textbf{0.7245} \\
    \midrule
    \multirow{4}{*}{PixelFormer} & Original & \multirow{4}{*}{\begin{tabular}{c}DIODE \\ Outdoor\end{tabular}} & 8.8726 & \underline{0.7041} & 1.4532 & \underline{21.8911} & \underline{66.0165} & 0.3254 & 0.5586 & \underline{0.7075} \\
    & Fine-tuned on KITTI & & 8.9302 & 0.7102 & \underline{1.4441} & 22.1350 & 66.4702 & 0.3244 & 0.5523 & 0.6929 \\
    & Fine-tuned on All Base &  & \underline{8.5296} & 0.7109 & 1.4768 & 22.0467 & 66.6546 & \underline{0.3270} & \underline{0.5531} & 0.7038 \\
    & Fine-tuned on All Enhanced & & \textbf{8.5109} & \textbf{0.7027} & \textbf{1.4408} & \textbf{21.5139} & \textbf{65.8426} & \textbf{0.3360} & \textbf{0.5635} & \textbf{0.7116} \\
    \bottomrule
  \end{tabular}
  \label{table:pf-diode-results}
\end{table*}

\begin{table*}
  \vspace{10pt}
  \caption{\textbf{Ablation Study: Monocular Depth Estimation performance of DPT-Hybrid fine-tuned on data from a model trained with no loss, a model conditioned on vanishing points with no loss and a model trained with our loss.} The best performing model is in bold.}
  \label{tab:ablation-table}
  \small
  \begin{tabular}{c|c|c|cccccccc}
    \toprule
    Model & Description & Test Set & RMSE $\downarrow$ & RMSE log $\downarrow$ & AbsRel $\downarrow$ & SqRel $\downarrow$ & SiLog $\downarrow$ & $\delta_1$ $\uparrow$ & $\delta_2$ $\uparrow$ & $\delta_3$ $\uparrow$ \\
    \midrule
    \multirow{3}{*}{DPT-Hybrid} & Fine-tuned on vKITTI No Loss & \multirow{3}{*}{KITTI} & 5.5733 & 0.2159 & 0.1573 & 1.1084 & 21.3919 & 0.7803 & 0.9389 & 0.9807 \\ 
    & Fine-tuned on vKITTI Conditioned & &  5.0437 & 0.1935 & 0.1402 & 0.8768 & 19.1673 & 0.8150 & 0.9499 & 0.9861 \\
    & Fine-tuned on vKITTI Enhanced &  &  \textbf{4.6749} & \textbf{0.1760} & \textbf{0.1250} & \textbf{0.7827} & \textbf{17.4836} & \textbf{0.8496} & \textbf{0.9608} & \textbf{0.9890}\\
    \midrule
    \multirow{3}{*}{DPT-Hybrid} & Fine-tuned on vKITTI No Loss & \multirow{3}{*}{\begin{tabular}{c}DIODE \\ Outdoor\end{tabular}} & 9.5241 & 0.5728 & 0.4573 & \textbf{6.7422} & 53.1904 & 0.4670 & 0.6581 & 0.7737 \\
    & Fine-tuned on vKITTI Conditioned & & 9.7312 & 0.5822 & 0.4641 & 7.1056 & 54.0504 & 0.4645 & 0.6520 & 0.7694 \\
    & Fine-tuned on vKITTI Enhanced & & \textbf{9.4854} & \textbf{0.5663} & \textbf{0.4559} & 6.8371 & \textbf{52.5902} & \textbf{0.4713} & \textbf{0.6595} & \textbf{0.7763} \\
    \midrule
    \multirow{3}{*}{PixelFormer} & Fine-tuned on vKITTI No Loss & \multirow{3}{*}{\begin{tabular}{c}DIODE \\ Outdoor\end{tabular}} & 8.5054 & 0.7047 & 1.3889 & 20.3750 & 66.5519 & 0.3184 & 0.5543 & 0.7035 \\  
    & Fine-tuned on vKITTI Conditioned & & 8.8021 & 0.7034 & 1.3923 & 19.4538 & 66.2341 & 0.3318 & 0.5592 & 0.7083 \\
    & Fine-tuned on vKITTI Enhanced & & \textbf{8.4728} & \textbf{0.6870} & \textbf{1.3738} & \textbf{19.3406} & \textbf{64.4721} & \textbf{0.3329} & \textbf{0.5677} & \textbf{0.7245} \\
    \bottomrule
  \end{tabular}
  \label{table:ablation-table}
\end{table*}

\begin{table}
  \vspace{10pt}
  \caption{\textbf{Inpainting Quantitative Results: Images generated by our enhanced model out-perform both the baseline Stable Diffusion V2 model and Ablations on the LPIPS metric.} Our enhanced model performs best on all three datasets, while the ablation model is outperformed by the baseline model when tested on only landscapes. Lower is better for all columns.}
  \label{tab:ablation-table}
  \small
  \begin{tabular}{c|c|c|c}
    \toprule
    Dataset & Holicity & Nature & All \\
    \midrule
    \# of Images & 250 & 320 & 570 \\
    \midrule
    Baseline & 0.1367 & 0.1584 & 0.1488 \\
    Ablation & 0.1147 & 0.1659 & 0.1434 \\
    Ours & 0.1138 & 0.1573 & 0.1382 \\
    \bottomrule
  \end{tabular}
  \label{table:inpainting-table}
\end{table}

\begin{table}
  \vspace{10pt}
  \caption{\textbf{FID Comparison: Images of non-building scenes generated by our enhanced model out-perform both the baseline Stable DiffusionV2 model and the No Loss model on the FID metric.} Metric was computed on 6.7k images from nature~\cite{landscapedata}, animal~\cite{animalsdata}, and indoor datasets~\cite{diode_dataset}. Lower is better.}
  \small
  \begin{tabular}{c|c|c|c}
    \toprule
    Model & Baseline & No Loss & Enhanced (Ours) \\
    \midrule
    FID $\downarrow$ & 23.1717 & 31.0726 & 21.1350 \\
    \bottomrule
  \end{tabular}
  \label{table:FID-table}
\end{table}

In order to evaluate the performance of our fine-tuned depth estimation models, we use both qualitative and quantitative measures. A qualitative comparison is shown in Fig.~\ref{fig:depth_qual}, while quantitative comparisons are in Table~\ref{tab:DPT-quant} and Table~\ref{tab:pixelformer-quant}.

\paragraph{DPT-Hybrid} We fine-tune one model from the base DPT-Hybrid using the generated vKITTI datasets and then test the model on both the original KITTI test set (Eigen Split) and a subset of the DIODE Outdoor test set. Results are in Table~\ref{tab:DPT-quant}. The models fine-tuned on images generated from our diffusion model outperform the original DPT-Hybrid model on all metrics on both datasets and outperform the model fine-tuned on images generated by the baseline model on all metrics for KITTI and all but one metric (SqRel) for DIODE Outdoor. In addition, for the DIODE Outdoor dataset, the original DPT-Hybrid model outperforms the base model on five out of eight metrics, but outperforms our model on no metrics. In particular, our model shows a 7.03\% improvement in RMSE and a 19.3\% improvement in SqRel over the original model while also demonstrating a 3.4\% improvement in SqRel and a 2.2\% improvement in SiLog over the baseline model. Fig.~\ref{fig:depth_qual} also shows qualitative comparisons between the original DPT-Hybrid model and the model fine-tuned on images generated by our enhanced diffusion model. Each set of images contains the input image, ground truth depth map (dilated with a 3$\times$3 kernel), and error maps from both the original model and our enhanced model. Additionally, the RMSE values for each of the depth predictions are shown in the top right of the error maps. The depth models from our model capture more high-frequency detail such as corners and poles, and also consistently have lower RMSE values.

\paragraph{PixelFormer} We fine-tune the base PixelFormer using both the generated vKITTI dataset and the full generated dataset and evaluate on the DIODE Outdoor test set. We additionally fine-tune a model using the original training set, KITTI~\cite{geiger2012are, agarwal2023attention}. Results are shown in Table~\ref{tab:pixelformer-quant}. The model fine-tuned on images from our diffusion model outperforms the original model, the models trained on images from the baseline model, and the model trained on KITTI on all metrics. Our model trained on the vKITTI dataset achieves a 4.1\% improvement in RMSE over the original model, while our model trained on the entire dataset achieves an 11.6\% improvement in SiLog over the original model and a 2.4\% improvement over the model trained on baseline images. Additionally, the original model outperforms the baseline model trained on the entire dataset on five of eight metrics, but outperforms the model trained on our images on no metrics.

\subsection{Human Subjective Tests}
\label{ssc:subj-tests-results}

Results from the human subjective tests are shown in Fig.~\ref{fig:subtest}. (a) shows the comparison between our enhanced model and the baseline model while (b) compares our enhanced model and the ablation model. Over all trials, images from our enhanced model appear more photo-realistic than images from the baseline model 69.6\% of the time and appear more photo-realistic than images from the ablation model 67.5\% of the time. In addition, the average rank of our images (between 1 and 3, lower is better) compared to the baseline was 1.9345 vs 2.4383 and was 1.9584 vs 2.4011 compared to the ablation model. The differences in average rank between our enhanced images and the baseline images (0.5038) and the difference between our images and the ablation images (0.4427) are also consistently less than the difference in average rank between our enhanced images and real images (0.3072 and 0.318 respectively). Overall, the results show that our proposed geometric constraint helps improve the photo-realism of generated images, as our enhanced images are consistently preferred over images from both the baseline model and ablation model.

\subsection{Ablation Study}
\label{ssc:ablation-results}

\begin{figure}
  \centering
  {\includegraphics[width=.45\textwidth]{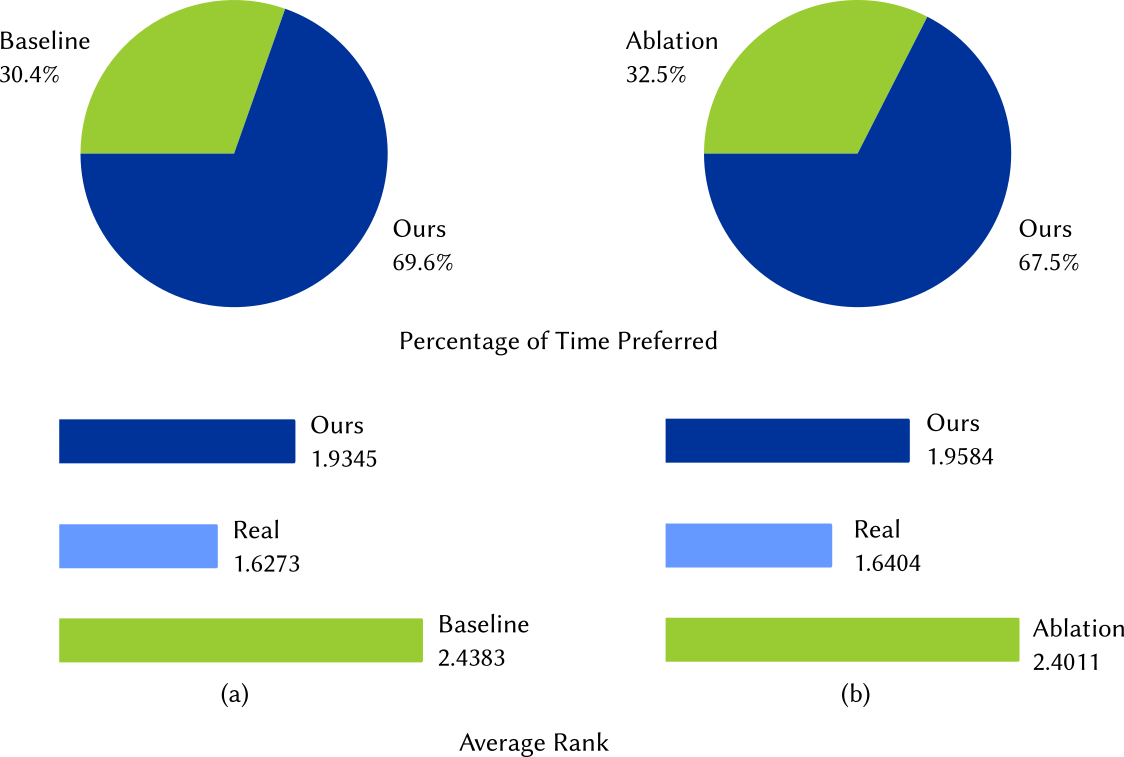}}
  \caption{\textbf{Images from our enhanced model consistently appear more photo-realistic than images from the baseline model (a) and our ablation model (b) according to the results of the subjective human tests.} \textit{Top.} How often each set of images was ranked lower. Our enhanced images were ranked as more photo-realistic (lower) than baseline images in 69.6\% of trials and were ranked as more photo-realistic than the ablation images in 67.5\% of trials. \textit{Bottom.} Average ranking for our images, real images, and comparison images. Although real images are consistently ranked the lowest, our images beat out both baseline and ablation images and are closer to real than the comparison.}
  \label{fig:subtest}
\end{figure}

To evaluate the value of our proposed constraint, we perform extensive comparison between our enhanced model and the ablation models. We include qualitative results comparing the no loss model and enhanced model in Fig.~\ref{fig:pespective_qual_ablation}. The edges and corners of our images are more consistent than similar features in the baseline model's images. We also include quantitative comparisons between depth estimation models trained on the vKITTI dataset from our enhanced diffusion model and depth estimation models trained on the vKITTI dataset from our no loss and conditioned diffusion models. The results from this experiment, for both DPT-Hybrid and PixelFormer, are shown in Table~\ref{tab:ablation-table}. The models trained on our enhanced model images outperform the models trained on the no loss model images on all metrics except for one (SqRel for DPT-Hybrid trained on the vKITTI dataset and tested on DIODE Outdoor). In addition, our model demonstrates significant improvements, up to 16.11\% on RMSE, compared to the no loss model. Our enhanced model also out-performs the conditioned model on all metrics. These results demonstrate that the superior performance of downstream models trained on our enhanced dataset is a result of our proposed constraint rather than a result of the new images introduced in fine-tuning. Beyond downstream tasks, the human subjective tests also show that our enhanced images are considered more photo-realistic than images from the no loss model 67.5\% of the time (Fig.~\ref{fig:subtest}). In addition, quantitative and qualitative results (Fig.~\ref{fig:inpainting_qual} and Table~\ref{table:inpainting-table}) on the inpainting task further highlight the improvement between our enhanced model and the no loss model. Combined, results from downstream tasks, human subjective tests, and the inpainting task demonstrate that the improvements achieved by our enhanced model are the result of our proposed geometric constraint rather than a result of fine-tuning on new images.

\section{Discussion}

\subsection{Limitations}
One of the key limitations of our approach is that fine-tuning the diffusion model requires a dataset of images with vanishing points during training. Although these can be approximated using vanishing point detection tools~\cite{lin2022deep, liu2021vapid}, these tools generally only work for images with strong vanishing lines. For images without these lines, such as nature scenes, our proposed loss would likely be ineffective. Another limitation of our approach is the generation speed of latent diffusion models. On average, it takes \textasciitilde3 seconds to generate a single image on 1 RTX3090, meaning generating a dataset of 150,000 images takes \textasciitilde125 hours on 1 RTX3090. This significantly limits the potential size of synthetic datasets generated by latent diffusion models. Another limitation is that although our images are improved compared to the baseline model's images, they are still not quite at the level of real images as shown by our subjective test results. For example, Fig.~\ref{fig:limitation} shows an image of Big Ben, and, although perspective lines are accurately depicted in the output, certain semantic details of the image are missing. Additionally, our technique only enforces perspective accuracy, meaning that other physical properties, such as lighting, shadows, or spatial relationships, may still be inaccurate.

\newcommand{\widthFigureLimitation}{.23\textwidth}
\begin{figure}
  \centering
  \captionsetup[subfloat]{labelformat=empty, skip=1pt}
{\includegraphics[width=\widthFigureLimitation]{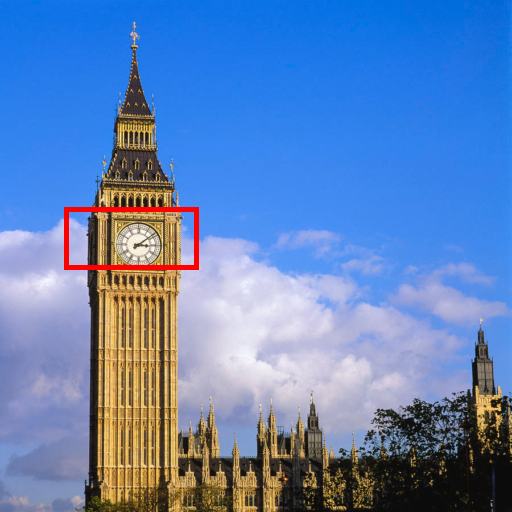}}
  \hfill
  {\includegraphics[width=\widthFigureLimitation]{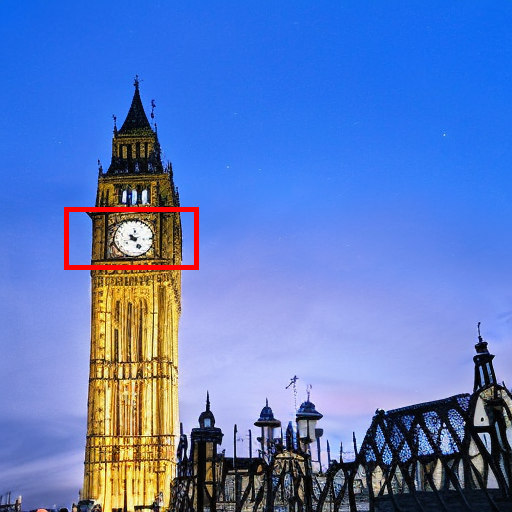}}  

    \vspace{1pt}
  
  \subfloat[\centering Real]
{\includegraphics[width=\widthFigureLimitation]{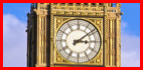}}
    \hfill
    \subfloat[\centering Ours Enhanced]{
    \includegraphics[width=\widthFigureLimitation]{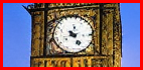}
    }
   \vspace{-10pt}
  \caption{\textbf{Outputs from stable diffusion are still unable to make certain semantic judgments.} Note that the clock shown on Big Ben is not functional and has no hour or minute hand.}
  \label{fig:limitation}
\end{figure}

\subsection{Societal Impact}

As always, there are downsides in improvements to generative models. As we increase the photo-realism of synthetic images, the potential for malicious use in the spread of disinformation also grows. In addition, perspective has been used as a tool to identify synthetic images from diffusion models~\cite{corvi2022detection}. With the addition of our constraint, these tools could lose their efficacy, further increasing the potential for misuse of diffusion models.

\subsection{Future Work}

The current work is limited to 3D geometry perspective constraints, but there are still many other physical properties that affect the realism of generated images. One such example is lighting and shadow consistency~\cite{farid_arxiv_perspective, farid_arxiv_lighting} and semantic and physical consistency. Images generated by diffusion models often break physical laws, for example by having people walking on water. Future work can explore other constraints to help fulfill these physical laws and further increase photo-realism and the performance of downstream tasks.

\subsection{Conclusions}
In the 1400s, Leon Alberta Battisti established the foundations for perspective in art, which pushed the boundaries of hand-drawn realism. In this work, we propose a first attempt at a novel geometric constraint which encodes perspective into latent diffusion models. We demonstrate that introducing these physically-based 3D perspective constraints improves both photo-realism on subjective tests and downstream performance on monocular depth estimation. We hope that our work can be a small step in our community effort to improve the realism of image synthesis.

\begin{acks}

We thank members of the Visual Machines Group (VMG) at UCLA for feedback and support, as well as Guha Balakrishnan and Krish Kabra for technical discussions. A.K. was supported by a \grantsponsor{0}{National Science Foundation (NSF)}{https://www.nsf.gov/} CAREER award \grantnum{0}{IIS-2046737}, \grantsponsor{1}{Army Young Investigator Program}{https://www.federalgrants.com/ARO-Young-Investigator-Program-YIP-13933.html} Award, and \grantsponsor{2}{Defense Advanced Research Projects Agency (DARPA)}{https://www.darpa.mil/} Young Faculty Award.

\end{acks}

\bibliographystyle{ACM-Reference-Format}
\bibliography{sample-base}

\end{document}